%% file: Main_journal.tex
\documentclass[10pt,journal,compsoc,twoside]{IEEEtran}
\usepackage{graphicx}
\usepackage{amsmath,amssymb,url}
\usepackage[usenames,dvipsnames]{xcolor}
\usepackage{float}
\usepackage{caption}
\usepackage{chemformula}
\usepackage{siunitx}
\usepackage{bm}
\usepackage{booktabs}

\ifCLASSOPTIONcompsoc
  \usepackage[nocompress]{cite}
\else
  \usepackage{cite}
\fi

\newif\ifcomments
\commentstrue

\ifcomments
    \usepackage[normalem]{ulem}

\else 
\fi

\def\Lair{L_{\rm air}}
\def\Lobj{L_{\rm obj}}
\def\Lobs{L_{\rm obs}}
\def\Lref{L_{\rm ref}}
\def\Tair{T_{\rm air}}
\def\That{\widehat{T}}
\def\Thatair{\widehat{T}_{\rm air}}
\def\emiss{\varepsilon}
\def\emisshat{\widehat{\varepsilon}}
\def\epsair{\varepsilon_{\rm air}}
\def\lambdaS{\lambda_{\rm sat}}
\def\Itilde{\widetilde{I}}
\def\Ldata{\mathcal{L}_{\rm DataFidelity}}
\def\Lreg{\mathcal{L}_{\rm Regularization}}
\def\yhat{\widehat{y}}
\def\dhat{\widehat{d}}
\def\dhatReg{\widehat{d}_{\rm Reg}}
\def\dhatNoReg{\widehat{d}_{\rm NoReg}}

\newcommand{\Frac}[2]{{{#1}/{#2}}}  

\usepackage{subfig}

\begin{document}

\title{Absorption-Based, Passive Range Imaging from Hyperspectral Thermal Measurements}

\author{Unay~Dorken~Gallastegi,
        Hoover~Rueda-Chac\'on,
        Martin~J.~Stevens,
        and~Vivek~K~Goyal
\IEEEcompsocitemizethanks{
\IEEEcompsocthanksitem This work was presented in part at the Conference on Lasers and Electro-Optics (CLEO), 2022~\cite{Gallastegi:22}. 
\IEEEcompsocthanksitem This work was supported in part by the US Defense Advanced Research Projects Agency (DARPA) Invisible Headlights program under contract number HR0011-20-S-0045
and in part by the US National Science Foundation under grant 1955219.
\IEEEcompsocthanksitem U. Dorken Gallastegi and V. K. Goyal are with the
Department of Electrical and Computer Engineering, Boston University,
Boston, MA, 02215 USA\@.
E-mail: \{udorken@bu.edu, v.goyal@ieee.org\}.
\IEEEcompsocthanksitem H. Rueda-Chac\'on was with Boston University at the initiation of this work.
He is with the Department of Computer Science, Universidad Industrial de Santander, Bucaramanga, 680002, Colombia.
E-mail: hfarueda@uis.edu.co.
\IEEEcompsocthanksitem M. J. Stevens is with the
National Institute of Standards and Technology, Boulder, CO, 80305 USA\@.
E-mail: martin.stevens@nist.gov.}
}

\markboth{Absorption-Based, Passive Range Imaging from Hyperspectral Thermal Measurements}%
{Dorken Gallastegi \MakeLowercase{\textit{et al.}}}

\IEEEtitleabstractindextext{%
\begin{abstract}
Passive hyperspectral longwave infrared measurements are remarkably informative about the surroundings.
Remote object material and temperature determine the spectrum of thermal radiance,
and range, air temperature, and gas concentrations determine how this spectrum is modified by propagation to the sensor.
We introduce a passive range imaging method based on computationally separating these phenomena.
Previous methods assume hot and highly emitting objects;
ranging is more challenging when objects’ temperatures do not deviate greatly from air temperature.
Our method jointly estimates range and intrinsic object properties, with explicit consideration of air emission, though reflected light is assumed negligible.
Inversion being underdetermined is mitigated by using a parametric model of atmospheric absorption and regularizing for smooth emissivity estimates.
To assess where our estimate is likely accurate, we introduce a technique to detect which scene pixels are significantly influenced by reflected downwelling.
Monte Carlo simulations demonstrate the importance of regularization, temperature differentials, and availability of many spectral bands.
We apply our method to longwave infrared (8--13\,\si{\micro\meter}) hyperspectral image data acquired from natural scenes with no active illumination.
Range features from 15\,\si{\meter} to 150\,\si{\meter}
are recovered, with good qualitative match to lidar data
for pixels classified as having negligible reflected downwelling.
\end{abstract}

}

\maketitle

\IEEEdisplaynontitleabstractindextext

\IEEEpeerreviewmaketitle


\IEEEraisesectionheading{\section{Introduction}\label{sec:introduction}}

\IEEEPARstart{R}{ange} imaging is valuable for a variety of applications such as autonomous navigation, inspection, topographical surveying, and mapping~\cite{9127841, bahnsen20213d, deems2013lidar, hubner2020evaluation}.
Methods that employ active illumination, such as lidar, can be very accurate but may not be suitable where stealth, power consumption, or eye safety are primary concerns \cite{royo2019overview, williams2017optimization}.
Stereo-based techniques are commonly used
for passive ranging \cite{9283161, matthies1994stochastic}. 
These techniques require precise pixel matching between a pair of images.
As a result, the performance degrades in dark and low-textured scenes, and for long-distance objects \cite{sibley2007bias, 5712095}. 

\begin{figure}
    \centering
        \includegraphics[width= .49\textwidth]{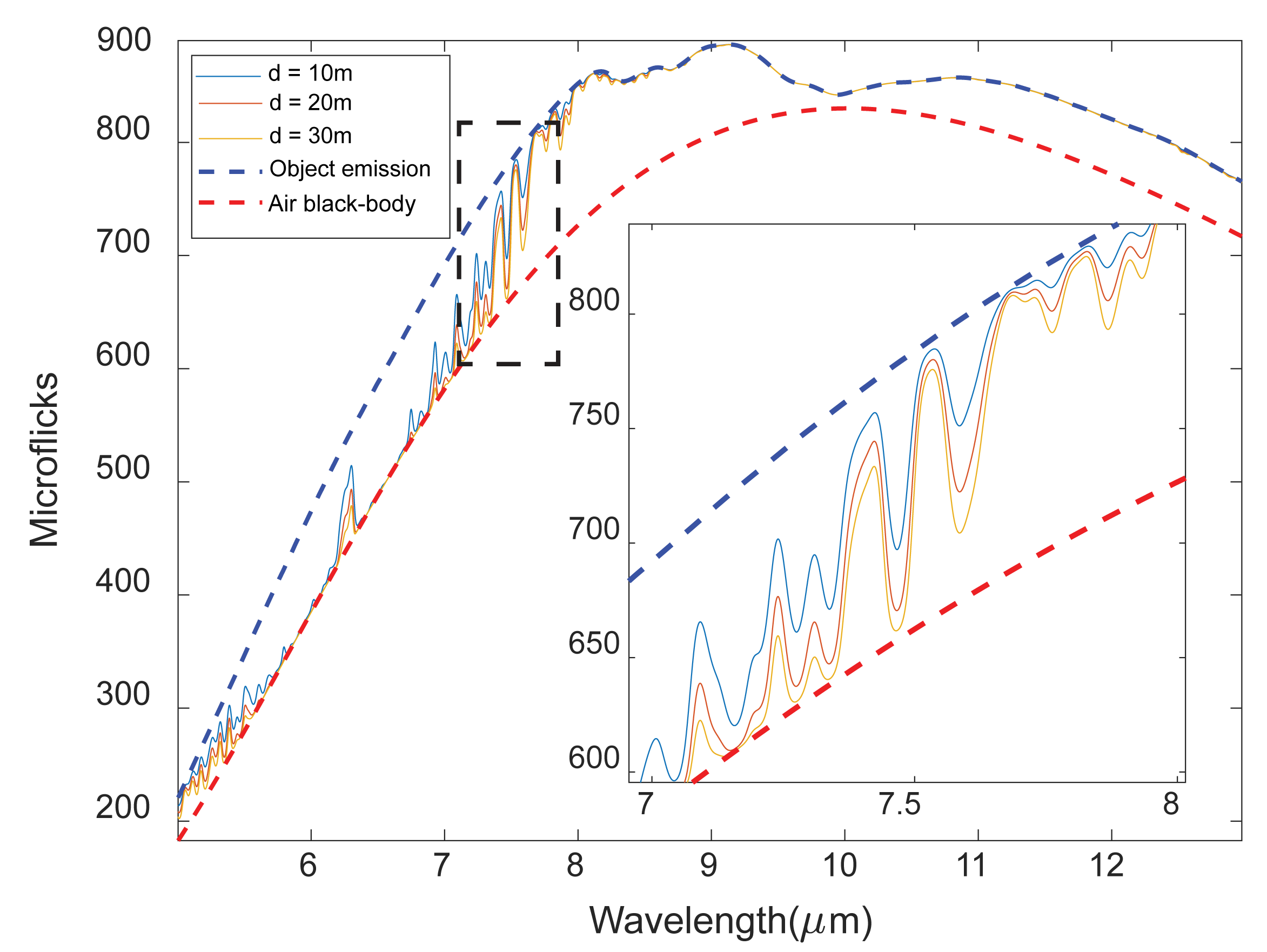}
    \caption{
    Simulation of hyperspectral measurements in units of microflicks (\si{\micro\watt \cdot sr^{-1} \cdot \centi\meter^{-2} \cdot \micro\meter^{-1}}) of a rock at 300\,\si{K} from 10\,\si{\meter}, 20\,\si{\meter}, and 30\,\si{\meter} ranges through the atmosphere at 289.7\,\si{K}. Nearly all the absorption in this range is due to water vapor, which here has a volume mixing ratio (VMR) of 0.012\@. The instrumental spectral response is a Gaussian with 40\,\si{\nano\meter} full width at half maximum. 
    Zoomed area of the spectrum is highlighted with a dashed box. 
    Atmospheric absorption depends on the wavelength and the range of the object.
    }
    \label{fig:Introduction1}
\end{figure}

\begin{figure*}
    \centering
    \subfloat[Depth map]{\includegraphics[width=.24\textwidth]{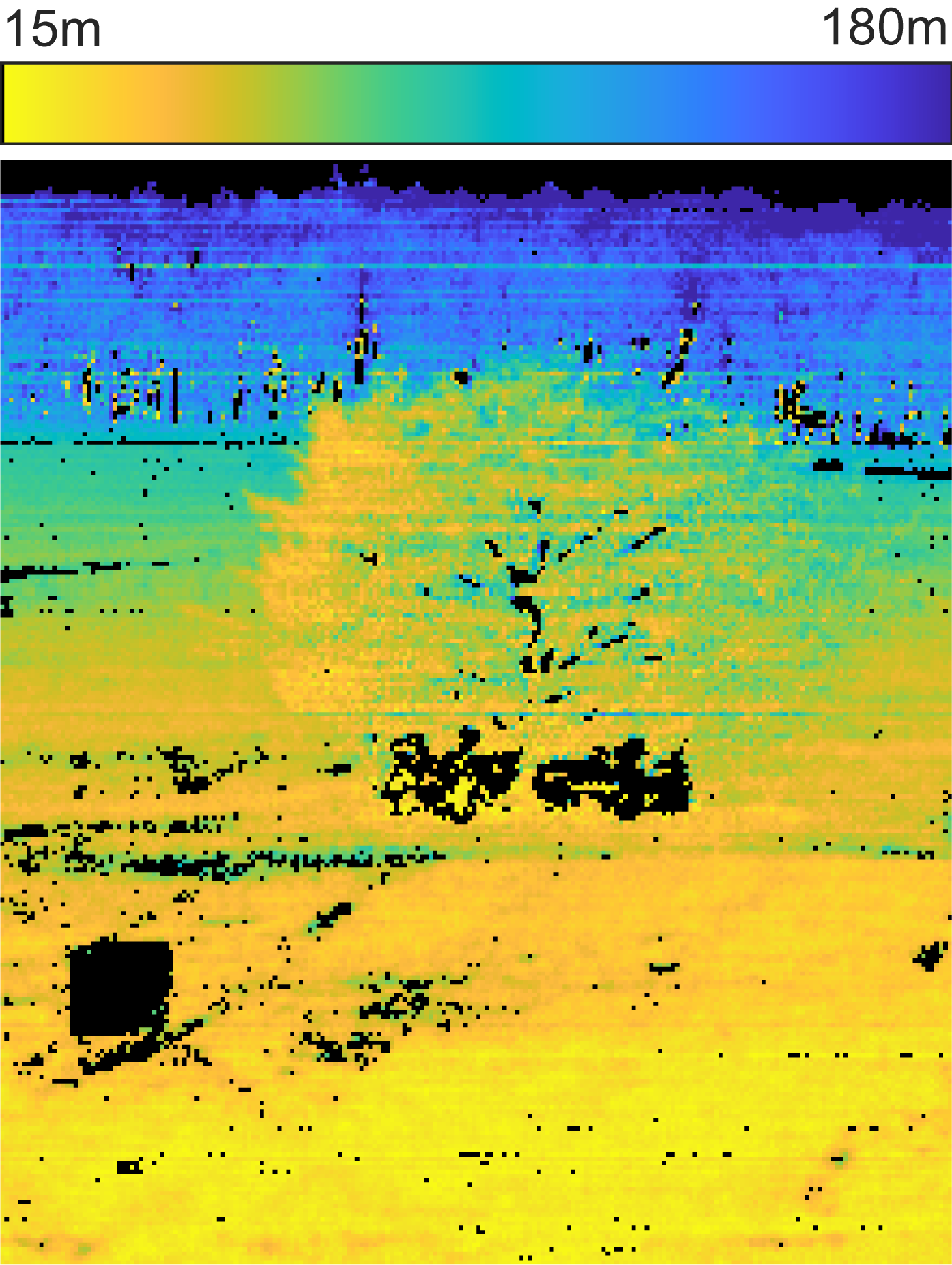}\label{fig:Teaser_Depth_Map}} \hfill
    \subfloat[Emissivity clusters]{\includegraphics[width=.24\textwidth]{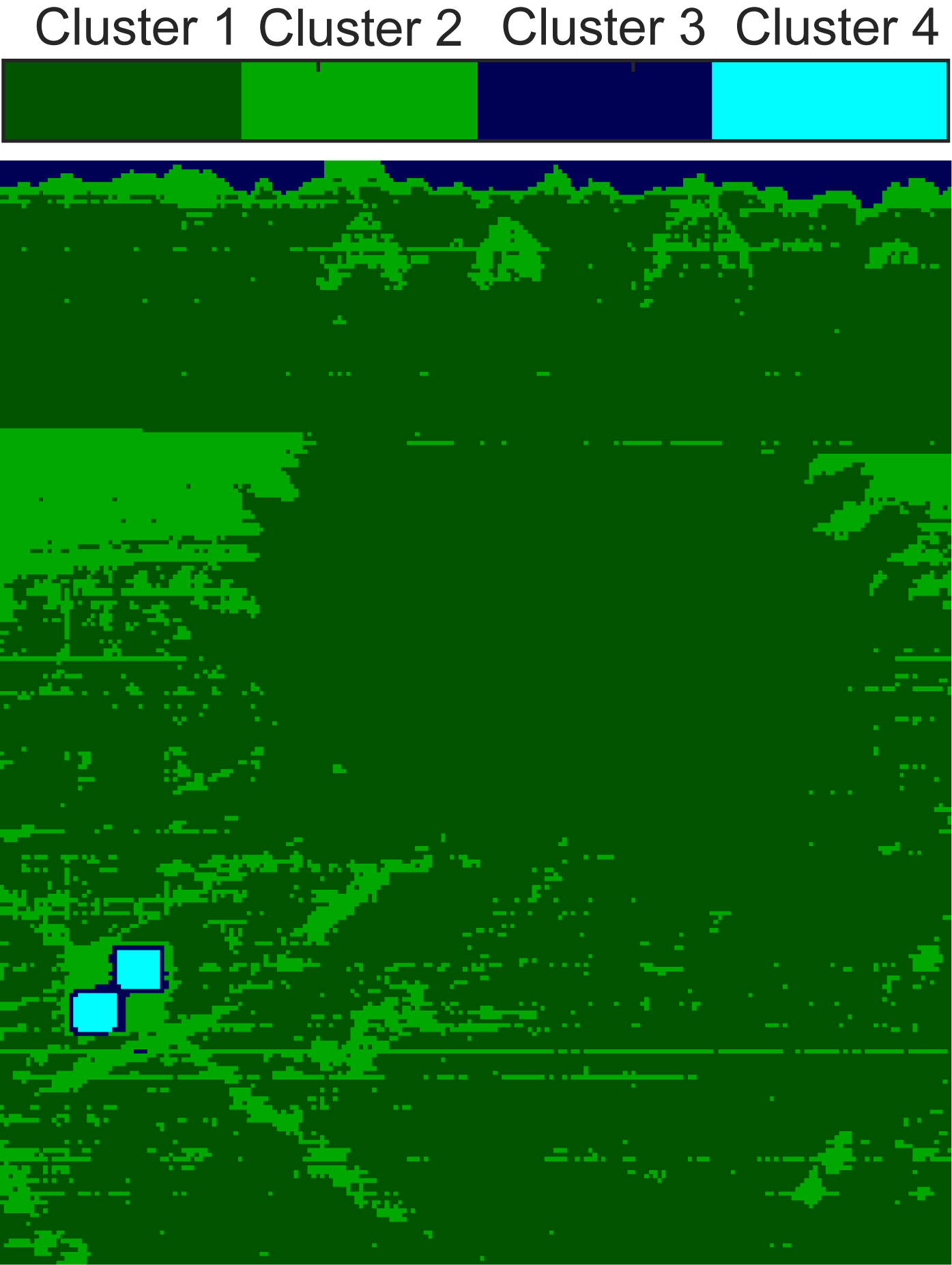}\label{fig:Teaser_Emissivity_Clusters}} \hfill
    \subfloat[Temperature map]{\includegraphics[width=.24\textwidth]{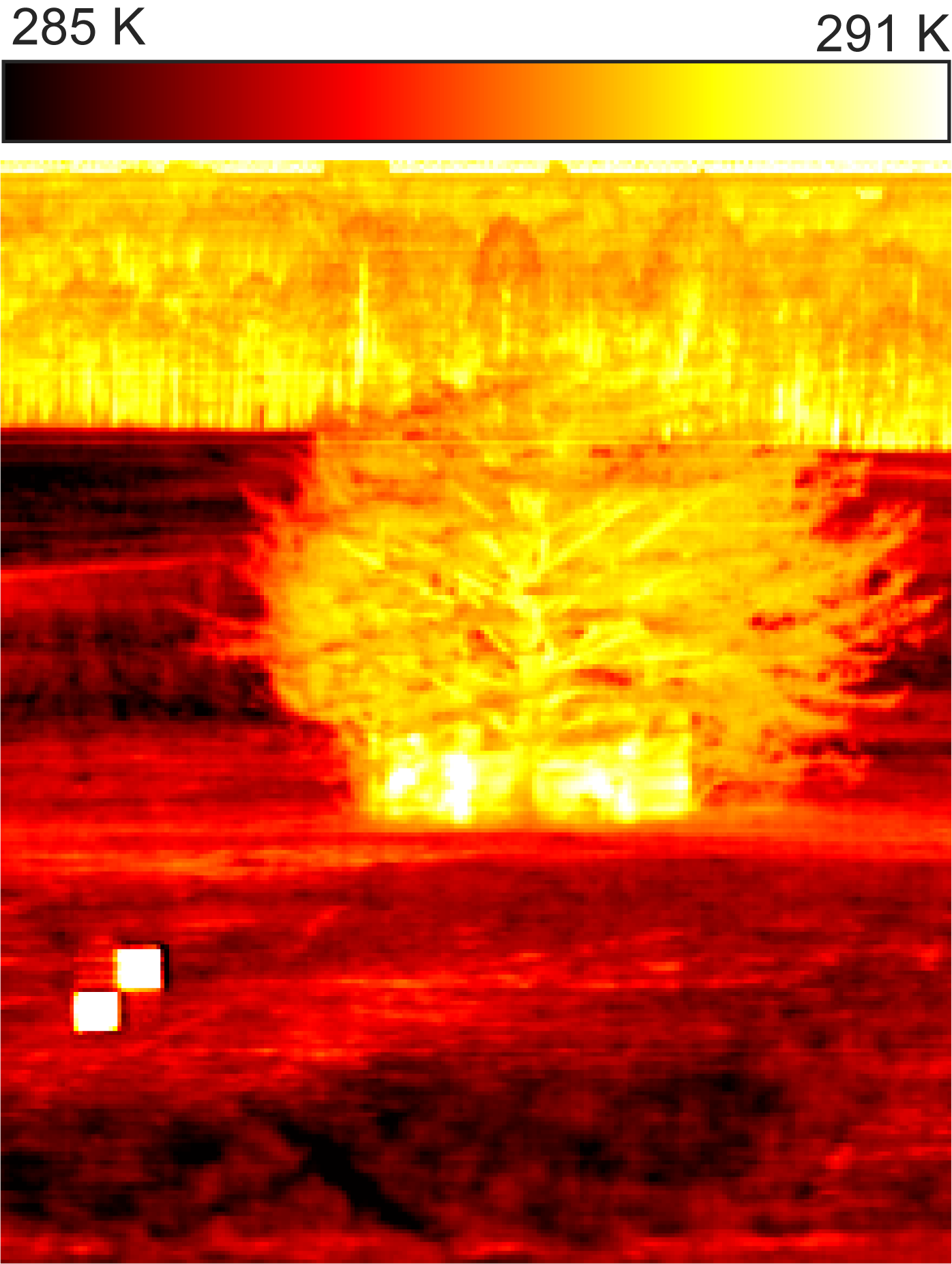}\label{fig:Teaser_Temperature_Map}} \hfill
    \subfloat[RGB image]{\includegraphics[width=.23\textwidth]{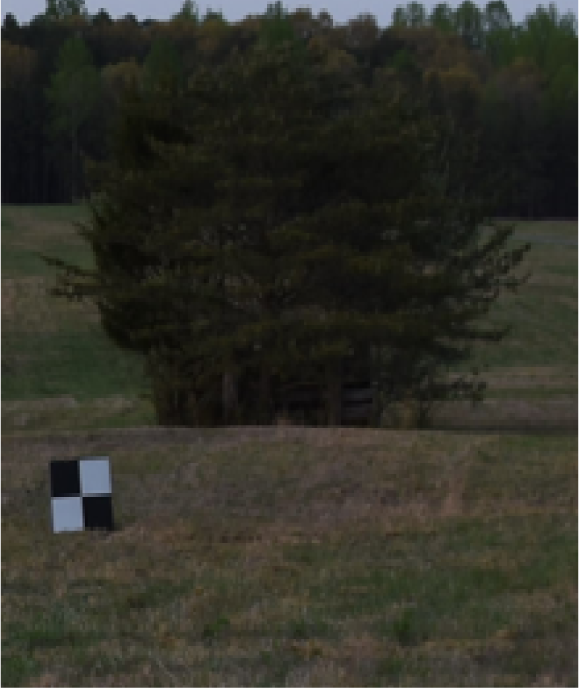}\label{fig:Teaser_RGB_Image}} 
    \caption{
    An example of the hyperspectral absorption-based ranging method of Section~\ref{sec:hyperspectral} on experimental data:
    \protect\subref{fig:Teaser_Depth_Map} depth map;
    \protect\subref{fig:Teaser_Emissivity_Clusters} emissivity clusters; and
    \protect\subref{fig:Teaser_Temperature_Map} temperature map.
    The range is extracted mostly from the absorptive spectral channels, with black pixels representing unreliable distance estimates due to significant downwelling contributions.
    Processing a wide spectral range enables object material identification.
    To show the value of the emissivity profile, $k$-means clustering is performed on the estimated emissivity profiles.
    The temperature is also estimated jointly, giving more information about the surroundings.
    The horizontal stripe artifacts in the results are due to data collection strategy with a scanning-based pushbroom sensor, as discussed in Section~\ref{sec:experimental_data_spec}.
    \protect\subref{fig:Teaser_RGB_Image} RGB image for visual reference of the scene.
    }
    \label{fig:Introduction}
\end{figure*}

Spectrally resolved measurements at atmospheric absorption bands have distance cues that can be exploited for ranging.
The key idea in absorption-based ranging is that air imparts a characteristic wavelength-dependent absorption profile on thermal radiance while it travels from an object to a sensor.
Fig.~\ref{fig:Introduction1} shows simulations of hyperspectral measurements of a rock observed at three different distances. 
The amount of radiance absorbed by the atmosphere is a function of distance and can be recovered by solving an inverse problem.
Recovering range from absorption requires measurements in at least two spectral channels~\cite{asano2020depth,kuo2021non, leonpacher1983passive, hawks2005passive}.

Previous works have demonstrated ranging of extremely hot objects such as fighter jet engines, missiles, or rockets \cite{leonpacher1983passive, hawks2005passive, macdonald2010passive, vincent2011passive, anderson2010monocular}. 
Absorption-based ranging is more challenging for natural scenes, where objects and air have similar temperatures.
Thermal emission from objects and air are then of similar magnitude,
and there are many scenarios where the air is more emissive than the surrounding objects~\cite{cermak2017eleven}.
Thus it is crucial to account for both atmospheric absorption and emission.
Furthermore, the low temperature contrast leads to higher uncertainty in range estimates.
Hyperspectral measurements can be used to reduce this uncertainty.
The absorption lines of the atmosphere are distributed over wide spectral windows.
For example, water vapor alone has thousands of absorption lines in the infrared.
Processing a wider band and extracting the distance information from a variety of absorption lines can dramatically improve range estimates from noisy data.

Since emissivities of the objects vary with wavelength,
the success of inversion depends on joint estimation of the emissivity and distance.
Emissivity estimation is important not only for accurate ranging but also for material identification \cite{mayer2002object, farrell2005impact, ozturk2015object}.
When hyperspectral measurements are available, prior spectral knowledge can be used to aid the inversion.
If air temperature, pressure and humidity are known, atmospheric absorption can be modeled accurately, exhibiting sharp features at specific, well-known wavelengths~\cite{GORDON2022107949, berk2014modtran}.
In contrast to the absorption lines of the atmosphere, thermal emission spectra from solid objects are typically smooth and continuous.
We exploit this physical phenomenon by using a regularization that promotes smooth emissivity estimates.

In this paper, we describe a passive 3D sensing method that can work in the dark and in natural conditions where temperature variations are low.
Alongside a depth map, our method also recovers the emissivity profiles and temperatures of objects in the scene.
We also provide a Fisher information analysis and demonstrate the performance of our method via simulations and through experimental results on data collected from natural scenes with a longwave infrared (LWIR) hyperspectral imager.
Fig.~\ref{fig:Introduction} shows an example output of our method on a hyperspectral data set with 256 spectral bands between 8\,\si{\micro\meter} and 13.2\,\si{\micro\meter}.
The stripe artifacts in the results are due to the datacube collection strategy in which a pushbroom sensor scans the scene horizontally, as discussed in Section~\ref{sec:experimental_data_spec}.
An RGB image of the scene is shown for reference but not used in the algorithm.

Our main contributions are:

\begin{itemize}
    \item Introduction of a measurement model including air temperature, air emission, and a parametric representation of the absorption spectrum of air.
    \item Joint estimation of range, emissivity, and temperature using regularization that promotes smooth emissivity estimates.
    We demonstrate the method on experimental hyperspectral data collected from 8\,\si{\micro\meter} to 13.2\,\si{\micro\meter}.
    \item Fisher information analysis, showing the influence of temperature variations in the scene, how range information is spread over the spectrum, and the 4.3\,\si{\dB} optimal attenuation level for estimating distance.
\end{itemize}

The main limitation of our method is that the small temperature difference between objects and the air in natural scenes results in noisy estimates.
Empirically, we associate the largest distance errors with significant reflected sky radiation contributions, otherwise known as downwelling radiance.
We introduce a method to detect large contributions from downwelling based on ozone absorption features, and we remove these pixels from our final result.
Available sensors typically avoid the most absorptive bands of the spectrum (6--8\,\si{\micro\meter}) and are thus not ideal for absorption-based ranging.
Nevertheless, our results show that the range can be recovered even with instruments that do not measure this band.
We would expect significantly better performance with future measurements that included these more absorptive parts of the spectrum.

After discussing related work in Section~\ref{sec:related},
we introduce a forward model including emission terms from the object and the air in Section~\ref{sec:forward}.
The Fisher information analysis in Section~\ref{sec:FisherInformationAnalysis} describes how the limits of ranging performance depend on temperature differentials and how information content is spread over the spectrum in a distance-dependent manner.
We present two approaches to distance estimation.
The bispectral method in Section~\ref{sec:bispectral} is based on a simple algebraic manipulation of the forward model
under the assumption that the object emissivities
are approximately equal at two nearby wavelengths.
To allow for objects close to air temperature,
in comparison to previous bispectral methods,
we include an air emission term,
and we show the importance of this term when analyzing experimental data.
Still, the low contrast in natural scenes results in poor range estimates when using only two wavelengths. 
Processing a wider band of the spectrum helps alleviate the effects of noise, but the emissivities of the objects are wavelength dependent and cannot be assumed constant over a wide band. 
In Section~\ref{sec:hyperspectral}
we introduce a ranging method based on solving a regularized optimization problem to separate the distance, emissivity, and temperature from hyperspectral measurements.
In Section~\ref{sec:simulation} we provide performance analyses of the methods with simulated data.
In Section~\ref{sec:experimental} we validate the methods with hyperspectral data collected from a natural scene.

\section{Related Work}
\label{sec:related}
Many prior publications have considered the effects of atmospheric absorption in spectral measurements.
Most work can be categorized into two types: methods that aim to recover range and methods that aim to remove the absorption effects to infer object emissivity. 
The latter can be useful for spectral classification or unmixing tasks.

\subsection{Absorption-Based Ranging}
Although our work focuses on passive methods, it is worth noting some prior studies that were performed using active illumination.
In \cite{asano2020depth}, 3D shapes of objects in water were recovered.
An incandescent light bulb was placed near the sensor,
and a depth map was estimated using the absorption as a result of a round-trip travel from the source to the sensor.
An extension of the method was proposed in \cite{kuo2021non}, where light scattering was also considered.
In \cite{kuo2021surface}, surface normals and a depth map were recovered by placing an illumination source at different angles.
These methods used either two or three spectral bands to extract information from the visible and near-infrared wavelengths between 400\,\si{\nano\meter} and 1\,\si{\micro\meter}.

Passive ranging has been performed using thermal radiation as a primary illumination source.
For example, passive ranging of hot objects has been demonstrated using atmospheric absorption.
One study from the 1980s focused on ranging a missile from its hot engine \cite{leonpacher1983passive} using the \ch{CO2} absorption band near 4.3\,\si{\micro\meter}. 
Another report~\cite{hawks2005passive} used the \ch{O2} absorption band at 765\,\si{\nano\meter}.
Both of these studies computed the ratio of radiance at an absorptive band to that at a nearby clear band, either measured or estimated.
This band ratio is an estimate of the atmospheric transmittance, which is mapped to range using a previously calculated transmittance--range curve.
Short-range~\cite{hawks2013short} and long-range~\cite{vincent2011passive} demonstrations of the method were shown by tracking a hot lamp and a Falcon 9 rocket, respectively.
Following the work on \ch{O2} absorption, a cheaper optical system was developed with three spectral filters~\cite{anderson2010monocular, yu2017passive}.
The system was later tested with a jet engine~\cite{anderson2011flight}.
A recent demonstration~\cite{nagase2022shape} showed ranging of hot objects around 323\,\si{K} to 363\,\si{K} using water vapor absorption around 8\,\si{\micro\meter}.

In the literature discussed so far, highly emitting objects such as missiles, rockets, jet engines, or light bulbs are analyzed; weaker emission sources are ignored and do not affect the accuracy much.
In natural scenes, by contrast, the temperature difference between objects and air is typically much smaller, often just a few degrees~\cite{cermak2017eleven}. 
In this case, all thermal emission sources should be considered, including the air.
Furthermore, the methods discussed so far use a small number of spectral channels.
For small temperature differences, measurements are much less sensitive to range, and it is beneficial to use as many spectral channels as possible.

In~\cite{DorkenGallastegi:24}, we explored the theoretical limits of absorption-based ranging in natural environments with low temperature contrast, utilizing representative emissivity profiles and various atmospheric models. That study included an analysis of different noise models, such as shot noise and readout noise. This paper extends that work by presenting different inversion methods and experimental validation.

\subsection{Atmospheric Compensation Techniques}

LWIR hyperspectral remote sensing has been used extensively to classify materials from long range~\cite{8738014}.
One of the major challenges in hyperspectral remote sensing is the atmospheric absorption effects in the collected measurements~\cite{borel2003artemiss}.
Atmospheric compensation techniques aim to remove the effects of the atmosphere from the measurements to infer object radiance.
Compared to the ranging methods for hot objects discussed in the previous section, these methods work for natural scenes, and the modeling of radiation includes an emission term from the air together with the objects.
The removal of atmospheric absorption effects is done by estimating the atmospheric transmittance function, which makes it highly related to our problem.
However, because all objects in the scene are approximately the same distance from the sensor, a single atmospheric transmittance function can typically be used for the entire scene.

In-scene atmospheric correction methods~\cite{young2002scene, borel2008error, boonmee2006land} extract the atmospheric transmittance function, which is then used to remove the absorption effects.
Recent work~\cite{kim2021at2es} extends the idea for close ranges, where transmittance is not assumed to be constant over the scene. 
In this case, knowledge of air temperature from a highly absorptive band helps to estimate the transmittance and the emissivity over patches of pixels.

The drawback of these in-scene correction methods, however, is that the estimated transmittance may not match the spectral structure expected of atmospheric gases because the estimation is done for each spectral band separately, without any prior information or constraint.
Model-based atmospheric compensation techniques, by contrast, recover the transmittance function from a predetermined lookup table~\cite{8738014}.
Although atmospheric compensation techniques are commonly used to infer object radiance, these are mainly developed to remove the atmospheric effects rather than using them for ranging.

\subsection{Thermal Hyperspectral Ranging}
Finally, we note that a recent paper
introduced a very different analysis method for data from the same collection:
heat-assisted detection and ranging (HADAR)~\cite{bao2023heat}.
HADAR maps infrared hyperspectral images to emissivity, temperature, and texture using a neural network.
In this decomposition, the texture component is taken to contain the range information.
Then, the emissivity, temperature and texture estimates are used to distill an RGB-like image to be used in previously developed monocular~\cite{masoumian2023gcndepth} or stereo~\cite{duggal2019deeppruner} range estimation techniques.
Although HADAR leverages hyperspectral measurements to separate emissivity, temperature, and reflected light components, the effects of atmospheric absorption and emission along the path from the object to the sensor are ignored.
Our work is thus complementary to HADAR\@. Future work combining aspects of HADAR with our forward model could lead to better estimates of range, temperature and emissivity than either approach alone.

\begin{figure*}
    \centering
    \includegraphics[width=1\textwidth]{Figure3/Figure3.png}
    \caption{Conceptual figure for radiative transfer model. Blue and red arrows represent the contributions from object emission and air emission to the observed spectrum, respectively.
    The gray arrow represent the reflected thermal radiation from other objects such as the sky and the rock.
    Nearly all the absorption features in the LWIR band (8--13\,\si{\micro\meter}) are due to water vapor.
    The attenuation coefficient of a model atmosphere with 0.012 volume mixing ratio of water vapor along with 289.7\,\si{K} temperature and 1010\,\si{\milli bar} pressure is plotted above.
    }
    \label{fig:forwardmodel}
\end{figure*}


\section{Forward Model of Atmospheric Absorption and Emission}
\label{sec:forward}
We use a standard radiative transfer model~\cite{manolakis2016hyperspectral}. The radiance arriving at the sensor is a combination of three terms, as illustrated in Fig.~3. First, each object in the scene emits thermal radiation. Second, each object reflects thermal radiation originating from its surroundings. The total radiance from emission and reflection reaching the sensor is attenuated by the wavelength-dependent atmospheric transmission function. Third, the atmosphere along the propagation path emits thermal radiation.

Starting from the object emission term, an ideal black-body at temperature $T$ emits light at wavelength $\lambda$ following Planck's law:
\begin{equation}
    B(\lambda; T) = \frac{2 \times 10^{-4}h c^2}{\lambda^5} \frac{1}{e^{h c/ \lambda k_\mathrm{B} T} - 1},
\end{equation}
where $h$ is the Planck constant, $c$ is the speed of light, and $k_\mathrm{B}$ is the Boltzmann constant~\cite{stewart2016blackbody}.
The radiance $B(\lambda; T)$ is typically expressed in units of microflicks (\si{\micro\watt \cdot sr^{-1} \cdot \centi\meter^{-2} \cdot \micro\meter^{-1}}).
Real-world materials emit less thermal radiation than true black-bodies.
The emissivity of an object, $\emiss(\lambda)$, is defined as the ratio between the radiance emitted by the object, $L(\lambda)$, and the radiance emitted by a black-body source at that temperature.
Therefore, the emission from a material with emissivity profile $\emiss(\lambda)$ and temperature $T$ can be expressed as
\begin{equation}
\label{eq:emissivity}
    L_{e}(\lambda) = \emiss(\lambda)B(\lambda;T).
\end{equation}

While light travels to the sensor through the atmosphere, some of it is absorbed.
The ratio between the radiance that reaches the sensor and the radiance emitted from an object is defined as the transmittance $\tau(\lambda;d)$. 
The contribution from the object to the observed radiance is then given by
\begin{equation}
\label{eq:ObjectContribution}
  \Lobj(\lambda) = \tau(\lambda;d)\emiss(\lambda)B(\lambda;T).
\end{equation}

The range information is embedded inside the transmittance function. 
Following the Beer--Lambert law, the transmittance for a distance $d$ can be expressed as
$$\tau(\lambda;d) = 10^{-\sigma(\lambda)\int_0^d  c(z) \, dz},$$
where $\sigma(\lambda)$ and $c(z)$ represent the molar attenuation coefficient and concentration along the path, respectively.
Assuming uniform concentration, this can be simplified to 
\begin{equation}
\label{eq:Trasnmittance}
    \tau(\lambda; d) = 10^{-\Frac{\alpha(\lambda) d}{10}},
\end{equation} 
where $\alpha(\lambda)$ is the wavelength-dependent attenuation coefficient, in units of \si{\dB/\meter}. For brevity, we call $\alpha(\lambda)$ the attenuation function.

In the visible part of the spectrum, scattering would contribute in three different components, including forward scattering, backward scattering, and exponentially attenuated direct component as discussed in~\cite{fujimura2018photometric, tsiotsios2017near}.
Scattering from the air is less pronounced in the LWIR band than it is in the visible and short-wave infrared bands~\cite{minnaert1995light, manolakis2016hyperspectral}.
In some cases, such as very foggy or dusty scenes, it may be important to consider scattering. 
The data analyzed here were collected in clear air conditions with little or no scattering. Therefore, we model the atmospheric transmittance only using absorption. 

For opaque objects, the reflectivity of the material can be related to its emissivity as $R(\lambda)=1 - \emiss(\lambda)$~\cite{manolakis2016hyperspectral}.
The incident radiance to the object can have contributions from many environmental sources, including  downwelling radiation from the atmosphere overhead (sky).
Assuming diffuse reflection from $N$ notable environmental sources as in~\cite{bao2023heat}, the contribution from the reflected thermal radiance can be modeled as
\begin{equation}
\label{eq:ReflectionContribution}
    \Lref(\lambda) = \tau(\lambda;d)(1-\emiss(\lambda))\sum_{i=1}^{N}\frac{\Omega_i}{\pi} L_{e,i}(\lambda),
\end{equation}
where $L_{e,i}(\lambda)$ and $\Omega_i$ represent the emission spectrum and the solid angle of the $i$th environmental object, respectively.
Atmospheric transmittance between the object and sensor is again represented as $\tau(\lambda;d)$.

We assume that light is either transmitted or absorbed by the air. 
Following Kirchhoff's law of thermal radiation, absorptivity is equal to emissivity under thermal equilibrium, and the emissivity of air can be related to transmittance as $ \epsair(\lambda) \approx 1 - \tau(\lambda;d)$. 
Although most natural scenes are not strictly at thermal equilibrium, this assumption is a good approximation when temperature variations are low~\cite{eismann2012hyperspectral}.
Following \eqref{eq:emissivity}, the path emission of the air at temperature $\Tair$ is
\begin{equation}
 \Lair(\lambda) = (1 - \tau(\lambda;d))B(\lambda;\Tair).
\end{equation}
In airborne or satellite-based remote sensing measurements, where the sensor is directed downward on a scene, this path emission is referred to as upwelling radiance~\cite{manolakis2016hyperspectral}.

Considering contributions from object emission, reflection, and atmospheric emission, the observed spectrum at the sensor can be expressed as
\begin{align}
 \Lobs&(\lambda) = \Lobj(\lambda) + \Lref(\lambda) + \Lair(\lambda) \nonumber \\
 &= \tau(\lambda;d)\emiss(\lambda)B(\lambda;T)  + \tau(\lambda;d)(1-\emiss(\lambda))\sum_{i=1}^{N}\frac{\Omega_i}{\pi} L_{e,i}(\lambda) \nonumber\\
 & \hspace{0.5cm} + (1 - \tau(\lambda;d))B(\lambda;\Tair).
\label{eq:ForwardModelTransmittance}
\end{align}
The visible part of the spectrum is typically dominated by reflected light from the sun and other light sources.
Thermal radiation emitted by natural objects is generally negligible compared to solar radiation at these wavelengths and is usually ignored.
In the LWIR band, by contrast,
reflected solar radiation usually has a negligible contribution, and thermal emission from scene objects and the atmosphere tend to dominate~\cite{8738014}.
This phenomenon causes a lack of contrast in the infrared images, often referred to as "ghosting"~\cite{bao2023heat, Gurton:14}.

In our analysis, we include the dominant object and atmospheric emission terms and neglect the reflected component.
This assumption holds especially well for natural objects with high emissivity, such as vegetation.
The forward model assuming only object and atmospheric emission simplifies to
\begin{align}
\Lobs(\lambda) &= \Lobj(\lambda) + \Lair(\lambda) \nonumber \\
 &= \tau(\lambda;d)\emiss(\lambda)B(\lambda;T) + (1 - \tau(\lambda;d))B(\lambda;\Tair).
\label{eq:ForwardModelTransmittance2}
\end{align}
We discuss the effect of neglecting the reflected component in Section~\ref{sec:results},
where we provide a method to identify pixels in the scene heavily affected by the reflected downwelling radiation,
which generally leads to overestimates of range.
These pixels are distinguished by the presence of the characteristic ozone absorption line around 9.6\,\si{\micro\meter} that is only present in the sky radiation. 

In the forward model in \eqref{eq:ForwardModelTransmittance2},
the observed spectrum is a convex combination of the object emission $\emiss(\lambda)B(\lambda;T)$ and a black-body at air temperature $B(\lambda; \Tair)$. 
Fig.~\ref{fig:Introduction1} shows how the observed spectrum is bounded between the object emission (blue dashed curve) and the black-body at air temperature (red dashed curve). 
Depending on the transmittance level, different bands of the spectrum provide different information. 
Around 6--7\,\si{\micro\meter}, the transmittance is very low ($\tau(\lambda)\approx 0$),
and the observed radiance is thus approximately equal to the air black-body curve;
this part of the spectrum can be used to find the air temperature. 
From 9--12\,\si{\micro\meter}, there are many bands where the transmittance is high ($\tau(\lambda)\approx 1$),
even for distances of 100\,\si{\meter} or more.
For these bands, the observed spectrum is close to the object emission, so it can be used to extract information about the intrinsic properties of the object, such as temperature and emissivity profile.
Other parts of the spectrum where the transmittance is neither very high nor very low contain distance information.

Rearranging terms in \eqref{eq:ForwardModelTransmittance2} and using the representation in \eqref{eq:Trasnmittance}, the observed radiance at the sensor is modeled as
\begin{equation}
\label{eq:ForwardModel}
    \Lobs(\lambda) = 10^{-\Frac{\alpha(\lambda) d}{10}}(\emiss(\lambda)B(\lambda;T) - B(\lambda; \Tair)) + B(\lambda; \Tair).
\end{equation}
Sensors typically have a finite spectral bandwidth, which impacts the discretized measurements they acquire.
Discretizing the continuous representation, a measurement of radiance at wavelength $\lambda_k$ can be written as
\begin{equation}
    y_k = \int_{0}^{\infty} f(\lambda - \lambda_k) \Lobs(\lambda) \, d\lambda, 
    \label{eq:Discretization}
\end{equation}
where $f(\cdot)$ is the instrumental spectral response function (ISRF)~\cite{beirle2017parameterizing}, which describes the sensor's wavelength-dependent response.
We approximated \eqref{eq:Discretization} by calculating the attenuation function with a high-resolution spectral modeling software package (SpectralCalc\footnote{Certain commercial products or company names are identified here to describe our study adequately. Such identification is not intended to imply recommendation or endorsement by the National Institute of Standards and Technology, nor is it intended to imply that the products or names identified are necessarily the best available for the purpose.})~\cite{SpectralCalc} that relies on high-resolution transmission molecular absorption database (HITRAN)~\cite{GORDON2022107949}, assuming a Gaussian ISRF with 40\,$\si{\nano\meter}$ full width at half maximum.
The software uses a standard model of the atmosphere~\cite{coesa1976standard}, modified to match the temperature, pressure and humidity values at the start of the data collect.
These values were collected with a weather station on site.
This yields an attenuation function $\alpha(\lambda)$ that we use to analyze all data.
The only atmospheric parameter varied in the analysis is $\Tair$ in \eqref{eq:ForwardModel}.


\section{Fisher Information Analysis}
\label{sec:FisherInformationAnalysis}

Next we analyze the Fisher information in each spectral channel.
We will use this analysis to show the limitations and optimal conditions for absorption-based ranging and how the depth information is spread over the spectrum.

We model the $k^{\text{th}}$
spectral measurement $y_k(d)$ of an object at range $d$ as a normal distribution
\begin{equation}
    y_k(d) \sim {\mathcal N}(\mu_k(d), \sigma^2),
\end{equation}
with mean $\mu_k(d)$ and variance $\sigma^2$.
The mean $\mu_k(d)$ is given by the forward model in \eqref{eq:ForwardModel}, and the variance $\sigma^2$ is assumed constant across the spectral channels.

Previously, we conducted an analysis under shot noise and readout noise modeling in~\cite{DorkenGallastegi:24}. Calculating the Fisher information under shot noise modeling requires knowledge of intrinsic camera parameters, such as integration time, pixel size, and solid angle. However, the publicly available experimental data does not include these parameters, making shot noise analysis infeasible. Consequently, we adopt a simpler noise model, additive white Gaussian noise (AWGN), for the analysis in this paper. If intrinsic camera parameters are available, Fisher information incorporating both shot noise and readout noise can be calculated using the framework outlined in~\cite{DorkenGallastegi:24}.

\begin{figure}
    \centering
    \subfloat[Simulated forward model at different relative temperatures.]
    {\includegraphics[width=.43\textwidth]{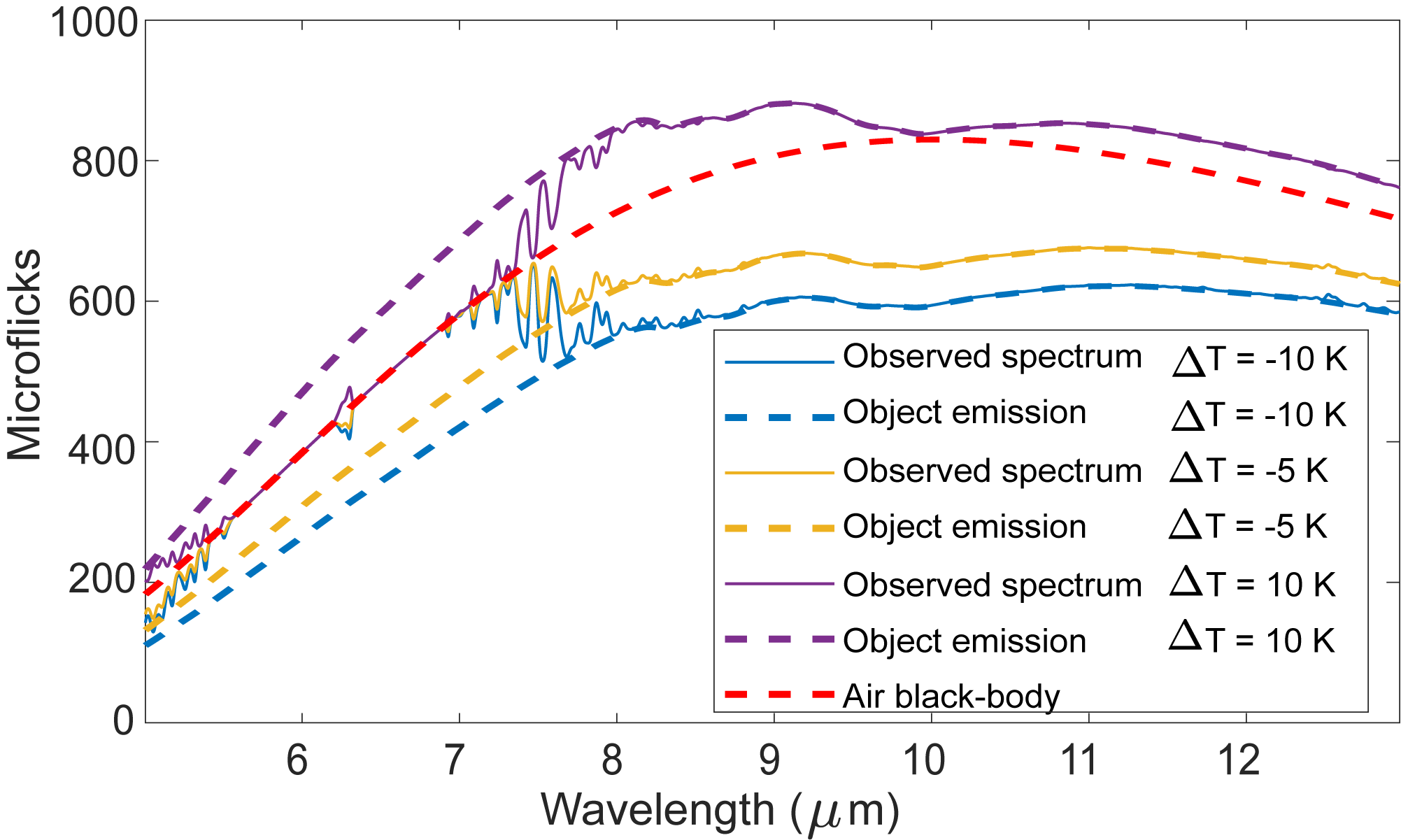}\label{fig:Fisher_information_Temperature_diff}} \\
    \subfloat[Fisher information analysis on spectrum.]
    {\includegraphics[width= .43\textwidth]{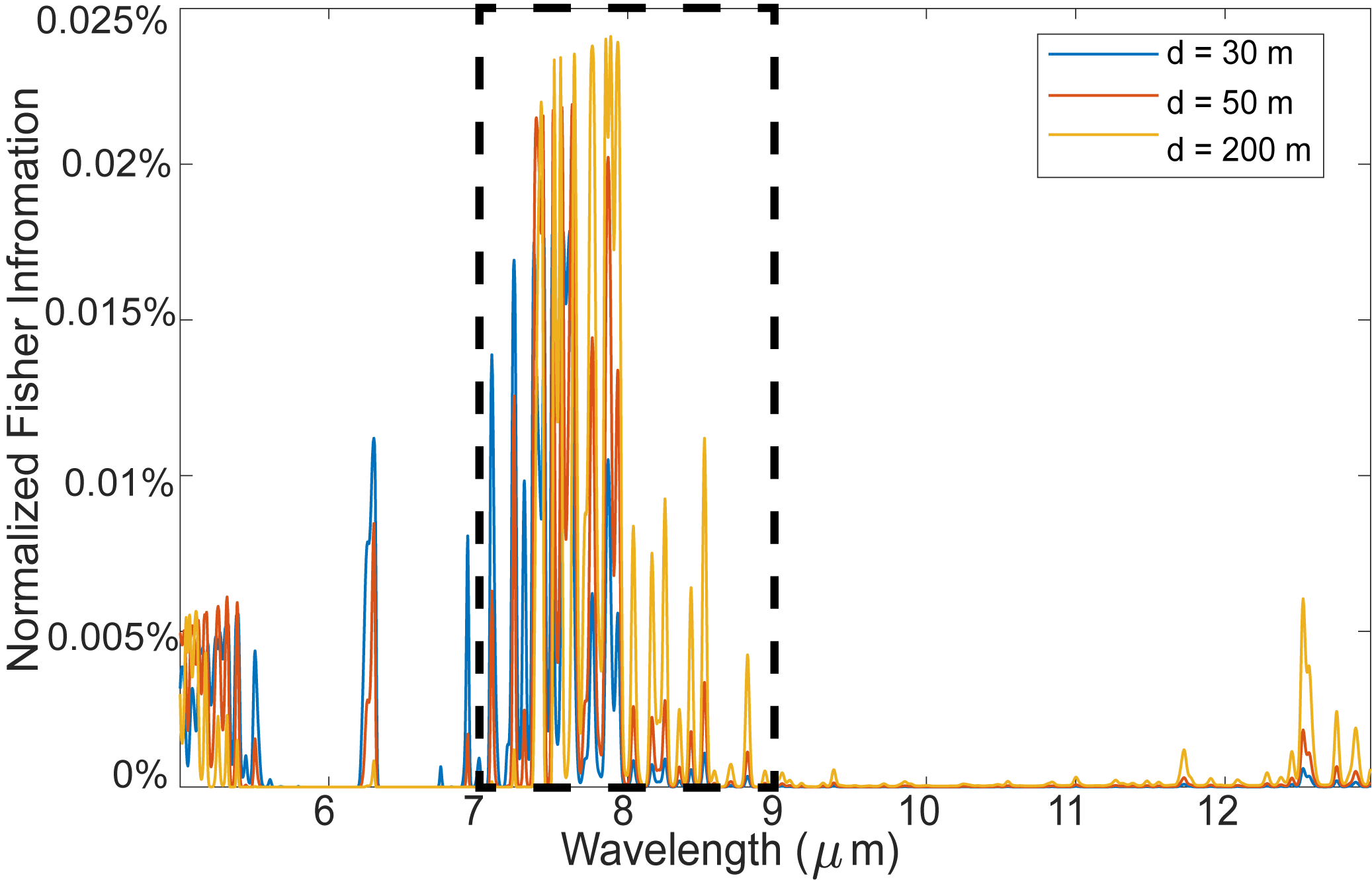}\label{fig:Fisher_information_Spectrum_1}} \\
    \subfloat[Zoom in on dashed box in (b).]
    {\includegraphics[width= .43\textwidth]{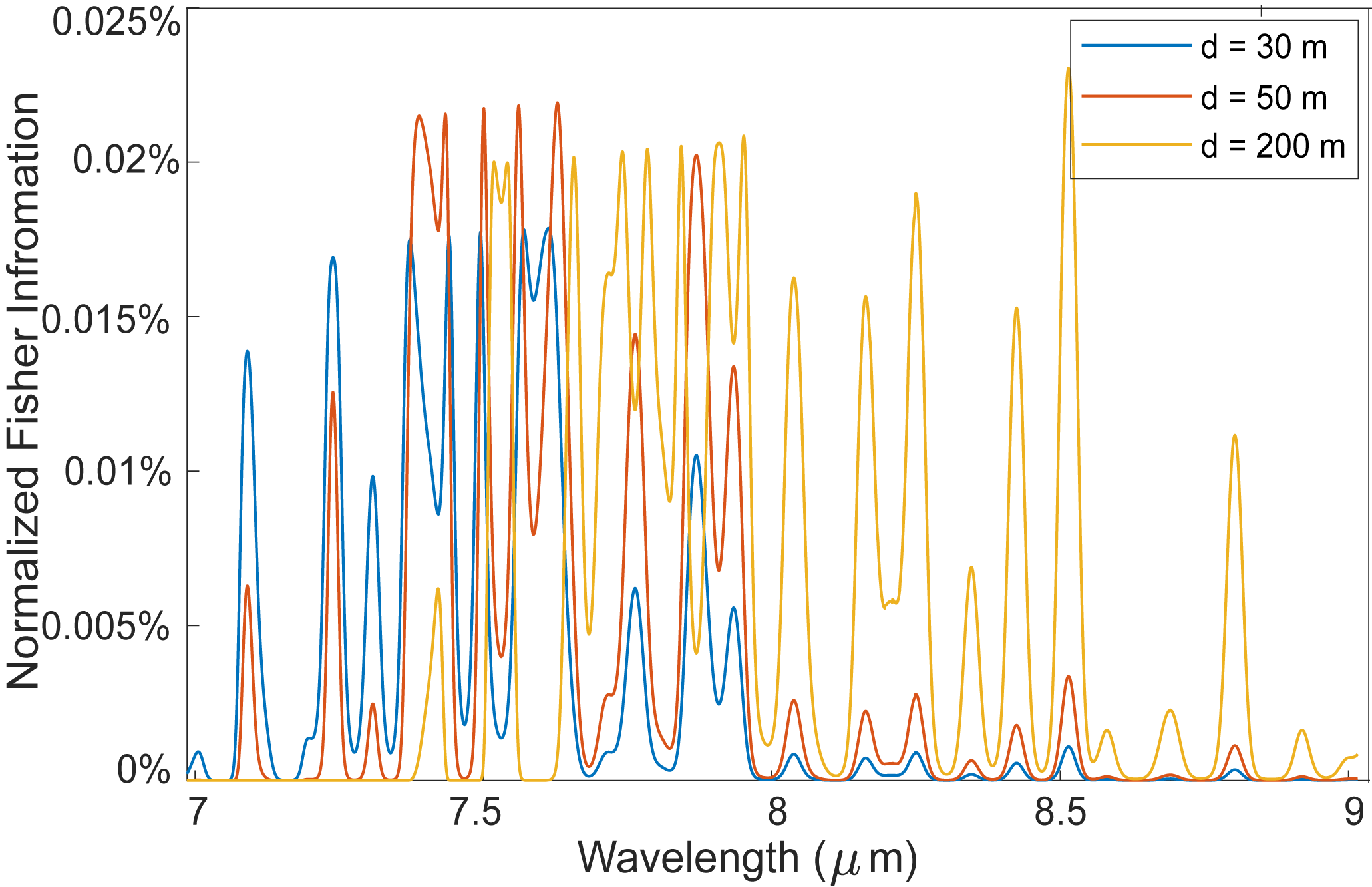}\label{fig:Fisher_infromation_Spectrum_2}}
    \caption{\protect\subref{fig:Fisher_information_Temperature_diff} Simulation of the forward model for a rock at \mbox{$-10$\,\si{K}}, $-5$\,\si{K}, and +10\,\si{K} relative temperatures with respect to air.
    The air temperature is 289.7\,\si{K}, and the corresponding black-body radiance is plotted with a dashed red line.
    The atmosphere is at 1010\,\si{\milli bar} pressure, and has 0.012 VMR water vapor content.
    The emissions from the object at different temperatures are plotted with dashed colored lines.
    Observed spectra at 30\,\si{\meter} range are shown in solid lines of matching colors.
    Objects with higher temperature differences show bigger atmospheric features.
    \protect\subref{fig:Fisher_information_Spectrum_1} Normalized Fisher information for a rock at 30\,\si{\meter}, 70\,\si{\meter}, and 200\,\si{\meter}.
    The information profile is dependent on the true distance and spread over the LWIR spectrum around the water vapor absorption lines.
    \protect\subref{fig:Fisher_infromation_Spectrum_2} shows the area in black box. For short distances, the most absorptive bands, in the 7--8\,\si{\micro\meter} region, contain the most information. As distance increases, information shifts to longer wavelengths. Most of the information in the collected measurement band (8--13\,\si{\micro\meter}) is in 8--9\,\si{\micro\meter}.
    }
    \label{fig:FisherInformation}
\end{figure}

The Fisher information for estimation of the mean radiance of spectral band $k$
is inversely proportional to its variance:
\begin{equation}
\label{eq:I_mu}
I_k(\mu_k(d)) = \frac{1}{\sigma^2}.   
\end{equation}
The Fisher information in spectral band $k$ for the distance parameter $d$ can be found by reparametrization of Fisher information as
\begin{equation}
I_k(d) =
  I_k(\mu_k(d))
  \left(\frac{\partial \mu_k}{\partial d}\right)^{\!2} .
\label{eq:FisherInformation}
\end{equation}
The partial derivative of the forward model in \eqref{eq:ForwardModel} with respect to distance is  given by
\begin{equation}
\frac{\partial \mu_k}{\partial d}
= -\frac{\ln(10)}{10}\alpha(\lambda_k)10^{-\Frac{\alpha(\lambda_k)d}{10}}(\emiss(\lambda)B(\lambda;T) - B(\lambda;\Tair)). 
\label{eq:partialderivative}
\end{equation}
This is proportional to the difference between the object emission and the emission of an ideal black-body at the air temperature, $\emiss(\lambda)B(\lambda;T) - B(\lambda, \Tair)$.
When this difference decreases, the measurements are less sensitive to distance, resulting in noisier estimates.
Fig.~\ref{fig:FisherInformation}(a) shows a simulation of the forward model of a rock at a 30\,\si{\meter} range at temperature differences of -5\,\si{K}, -10\,\si{K}, and +10\,\si{K} with respect to the air temperature.

\begin{figure*}
    \centering
    \subfloat[Neglecting air emission]{\includegraphics[width=.24\textwidth]{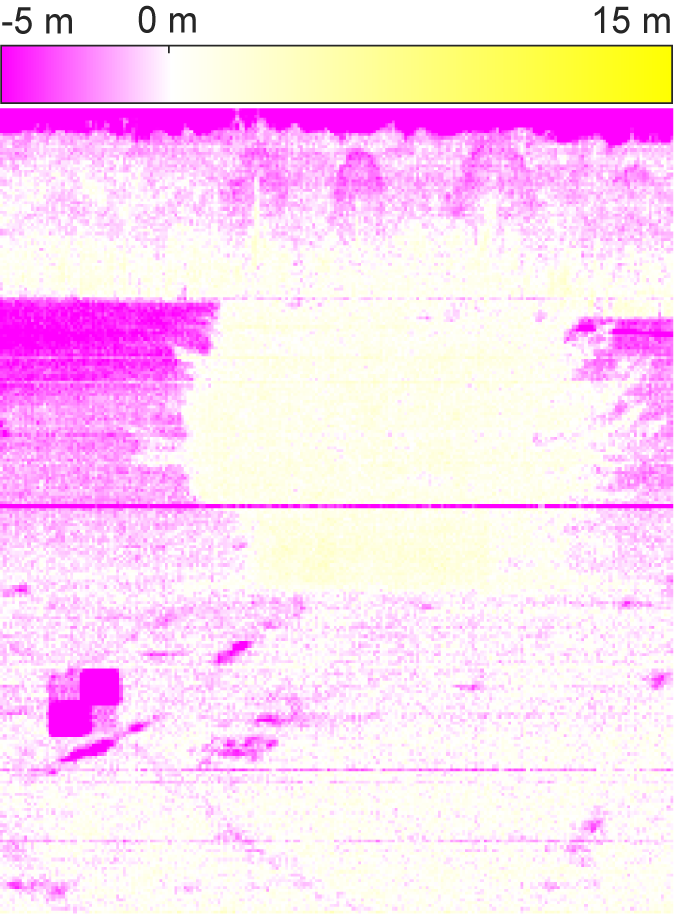}\label{fig:Air_emission_Neglecting_air}} \hfill
    \subfloat[Including air emission]{\includegraphics[width=.24\textwidth]{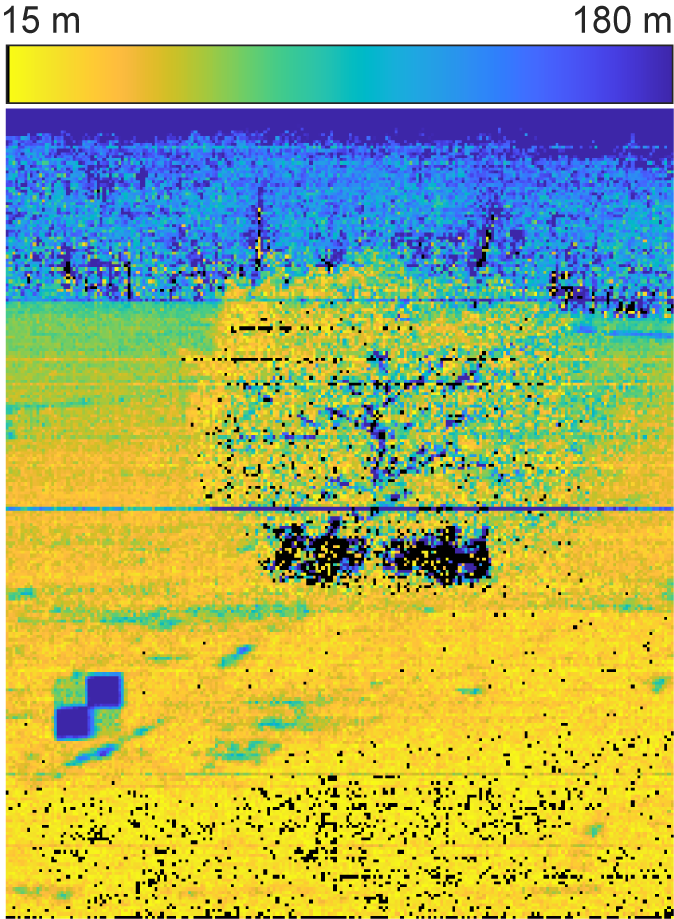}\label{fig:Air_emission_Including_air}} \hfill
    \subfloat[Lidar]{\includegraphics[width=.24\textwidth]{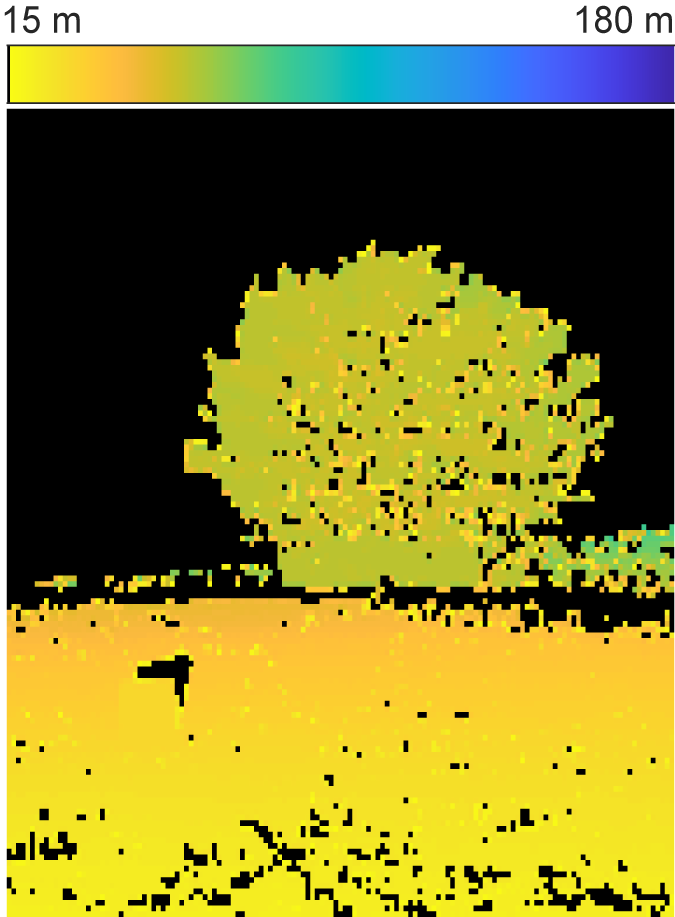}\label{fig:Air_emission_lidar}} \hfill
    \subfloat[Temperature difference]{\includegraphics[width=.24\textwidth]{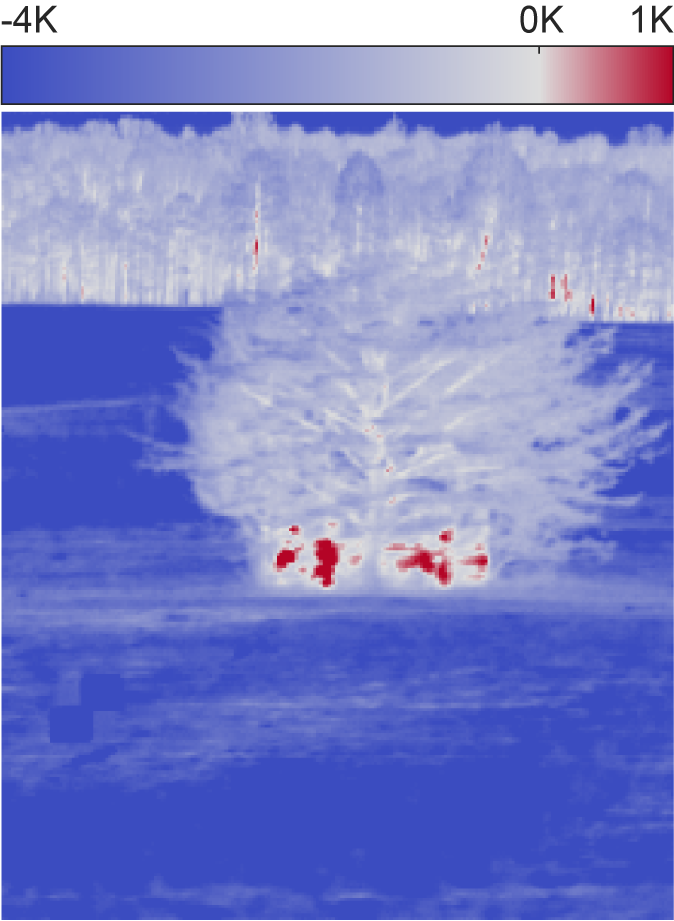}\label{fig:Air_emission_Temperature_difference}} \hfill
    \caption{Bispectral ranging using one clear (8.38\,\si{\micro\meter}) and one absorptive band (8.42\,\si{\micro\meter}).
    Depth maps for
    \protect\subref{fig:Air_emission_Neglecting_air} neglecting and
    \protect\subref{fig:Air_emission_Including_air} including the air emission.
    \protect\subref{fig:Air_emission_lidar} Depth map produced by lidar for comparison,
    without pixel-to-pixel correspondence with the hyperspectral sensor;
    black pixels represent missing data.
    Note that the color bar range is different for the first plot because most of the range estimates are negative or close to zero when air emission is neglected.
    \protect\subref{fig:Air_emission_Temperature_difference} Estimated temperature difference with respect to air, calculated using brightness temperature at the clear band for object temperature and weather station data collected on-site for air temperature.} 
    \label{fig:Bispectral_Ranging2}
\end{figure*}

\begin{figure*}
    \centering
    \subfloat[Temperature difference and depth maps of the scene over 90 minutes.]
    {\includegraphics[width=1\textwidth]{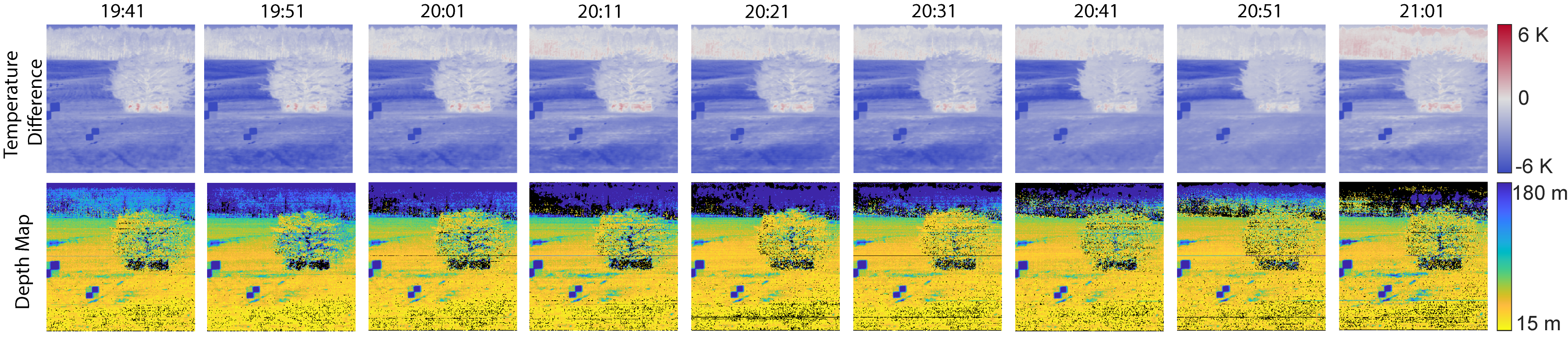}\label{fig:DepthRecBispectral}} \\
    \subfloat[Temperature profile of the scene over 90 minutes]
    {\includegraphics[width=1\textwidth]{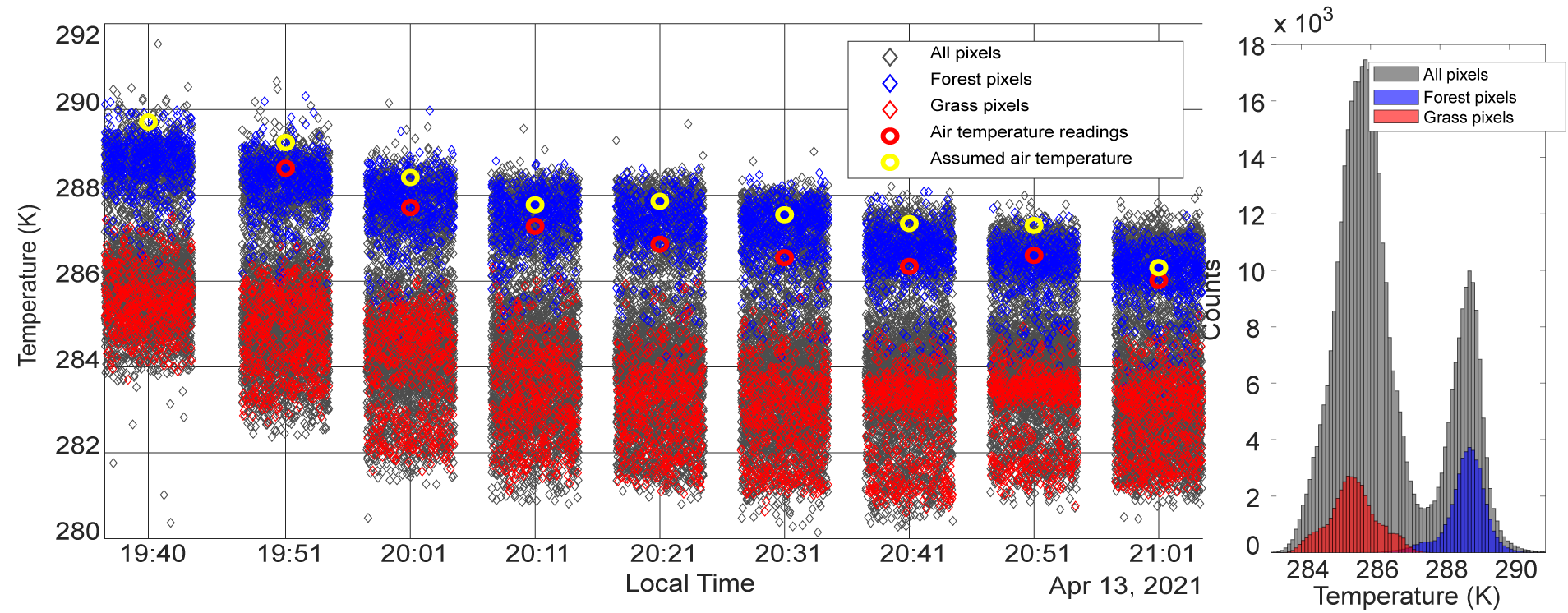}\label{fig:TemperatureProfile}}
    \caption{Bispectral ranging results on experimental data.
    \protect\subref{fig:DepthRecBispectral} Bispectral ranging of the scene over 90 minutes, with local time passing from 19:41 to 21:01 in steps of 10 minutes.
    The temperature difference with respect to assumed air temperature is shown in the top row and depth maps are shown in bottom row. 
    \protect\subref{fig:TemperatureProfile} Temperature profile of the scene as a function of time.
    Temperatures of each pixel are plotted with gray diamonds, uniformly spread over 7 minutes randomly for each acquisition for visualization purposes.
    There are two main thermal regions in the scene:
    the background forest (blue diamonds), and
    the foreground grass (red diamonds), both being mostly cooler than the air.
    The air temperature readings from the weather station are shown with red circles.
    For results in~\protect\subref{fig:DepthRecBispectral}, where artifacts from wrong air temperature are visually minimized, the assumed air temperature is plotted in yellow circles;
    these deviate at most 1\,\si{K} from the raw air temperature readings.
    At 19:41, assumed and recorded air temperature are the same.
    In an ideal case, the air temperature should be extracted from a saturated band using brightness temperature for each pixel.
    For the first acquisition, temperatures are also shown as a histogram with two prominent peaks.
    The figure highlights the effectiveness of bispectral ranging in recovering depth maps across varying temperature distributions over time.
    }
    \label{fig:Bispectral_ranging_time}
\end{figure*}

When there is a large difference between object and air temperatures (the purple and blue dashed curves) measurements will exhibit larger atmospheric absorption features and higher Fisher information compared to when the temperature difference is small (yellow dashed curve).
For the extreme case when the object emission is equal to the air black-body, the measurements are not sensitive to range and the Fisher information is zero.

Assuming the spectral measurements are independent, the total Fisher information for $K$ channels can be written
\begin{equation}
I(d) = \sum_{i = 1}^K I_k(d) = \frac{1}{\sigma^2}\sum_{k = 1}^K \left(\frac{\partial \mu_k}{\partial d}\right)^{\!2} .
\end{equation}
Next, we define the normalized Fisher information $\Itilde_k$ as the ratio between the Fisher information in channel $k$, $I_k(d)$, and the total Fisher information in the spectrum $I(d)$:
\begin{equation}
    \Itilde_k(d) = \frac{I_k(d)}{I(d)} = \frac{\left(\frac{\partial \mu_k}{\partial d}\right)^{\!2}}{\sum_{k=1}^K \left(\frac{\partial \mu_k}{\partial d}\right)^{\!2}} .
\end{equation}

Fig.~\ref{fig:FisherInformation}(b) shows the normalized Fisher information of spectral channels for a 284.7\,\si{K} rock at ranges of 30\,\si{\meter}, 50\,\si{\meter}, and 200\,\si{\meter}, with 289.7\,\si{K} air temperature and 40\,\si{\nano\meter} ISRF.
The distance information lies in partially absorptive channels that are spread over the spectrum.
The spread of the information along the spectrum depends on the true target range.
For a nearby object (blue curve for 30\,\si{\meter}) the information is concentrated on very absorptive bands (7--8\,\si{\micro\meter}) where there is significant radiance from ambient-temperature objects.
For a more distant object (yellow curve for 200\,\si{\meter}), the normalized information in highly absorptive channels decreases and the information in less absorptive channels increases.
This behavior is expected since at highly absorptive bands the radiance emitted from the object is attenuated very quickly with distance and does not reach the sensor.
This leads to a trade-off in the ideal attenuation level.
If there is no attenuation, distance cannot be estimated since the measurements are not sensitive to range. On the other hand, if there is too much attenuation, the emitted radiance does not reach the sensor and the measurements are again insensitive to range.

There is an optimal attenuation level for absorption-based ranging, dependent on the true target distance. 
This can be derived by using gradient analysis to maximize the Fisher information.
The resulting optimal attenuation coefficient $\alpha^{\ast}(d)$ for a distance $d$ is
\begin{equation}
    \alpha^{\ast}(d) = \frac{10\,\si{\dB}}{\ln(10)d}.
\end{equation}
This results in attenuation
\begin{equation}
    \alpha^{\ast}(d)d
    \approx 4.3\,\si{\dB}
\end{equation}
or $1/e$ on a linear scale.
To maximize the information obtained for an object at a range $d$, one should choose a wavelength $\lambda$ such that $\alpha(\lambda)=\alpha^{\ast}(d)$.

Focusing on the LWIR part of the spectrum, most of the band is close to transparent and not ideal for close ranges.
There are low absorption channels around 8--9\,\si{\micro\meter} and 12--13\,\si{\micro\meter} that may be suitable for ranging objects over 200\,\si{\meter} away from the sensor.
The attenuation function of the atmosphere is not controllable by the user, but it shows a variety of attenuation levels when we consider a wide spectral window.
Hence, it is best to use many spectral channels over a wide spectral bandwidth to range objects at a variety of distances.
Hyperspectral imaging facilitates data acquisition across wavelength bands optimized for a wide range of distances.
\section{Bispectral Ranging}
\label{sec:bispectral}
Estimating range using two spectral channels has been demonstrated before in absorbing media such as water~\cite{asano2020depth,kuo2021non} or air~\cite{leonpacher1983passive, hawks2005passive}. 
Previous ranging methods have ignored thermal emission from air and therefore have simply inverted \eqref{eq:ObjectContribution}.
For ranging to be accurate in natural scenes, it is crucial to account for air emission, and therefore, \eqref{eq:ForwardModel} should be inverted instead. 
For this, the air temperature is assumed to be known, either from another sensor or estimated from a saturated absorptive band ($\lambdaS$), where the transmittance is close to zero ($\tau(\lambdaS) \approx 0$), via
\begin{equation}
    \Thatair = B^{-1}(\lambdaS; \Lobs(\lambdaS)),
\end{equation}
where $B^{-1}$ is the inverse of the black-body function, otherwise known as brightness temperature~\cite{manolakis2016hyperspectral}. 
At two absorptive bands ($\lambda_1$ and $\lambda_2$), the measurements are
\begin{align}
    \Lobs(\lambda_i) = 10^{-\Frac{\alpha(\lambda_i)d}{10}}(\emiss(\lambda_i)B(\lambda_i, T) - B(\lambda_i,\Tair)) \nonumber\\
     + B(\lambda_i; \Tair), \qquad i = 1,2.
\end{align}
Similarly to other methods working in a narrow spectral window, we will assume that the emission factors are the same at nearby spectral bands, that is,
\begin{equation}
    \emiss(\lambda_1)B(\lambda_1;T) - B(\lambda_1;\Tair) \approx \emiss(\lambda_2)B(\lambda_2;T) - B(\lambda_2;\Tair).
    \label{eq:SameEmissivity}
\end{equation}
With these assumptions, an estimator for range can be computed as
\begin{equation}
    \dhat = \frac{-10}{\alpha(\lambda_1) - \alpha(\lambda_2)}\log_{10}\!\left(\frac{\Lobs(\lambda_1) - B(\lambda_1;\Thatair)}{\Lobs(\lambda_2) - B(\lambda_2;\Thatair)} \right).
\end{equation}
For spectral channels farther apart, the assumption in \eqref{eq:SameEmissivity} may not hold.
In this case, a third spectral measurement is beneficial so that the object emission at the absorptive band can be estimated from two nearby clear bands \cite{anderson2010monocular}.
It should be noted that there is not enough information in a two-wavelength measurement for a joint estimation of distance, temperature and emissivity.
However, once distance has been estimated, the object radiance term~\eqref{eq:emissivity} can be recovered as
\begin{equation}
    \widehat{L_e}(\lambda_i) = \frac{L_{obs}(\lambda_i) - B(\lambda_i; T_{air})}{10^{-\alpha(\lambda_i)\hat{d}/10}} + B(\lambda_i; \Tair), \qquad i = 1,2.
\end{equation}

Fig.~\ref{fig:Bispectral_Ranging2} shows range estimation results from experimental data using only one absorptive band ($\lambda_1$ = 8.42\,\si{\micro\meter}, $\alpha(\lambda_1) = 8.6 \times 10^{-4}$\,\si{\dB/\meter}) and one transparent band ($\lambda_2$ =  8.38\,\si{\micro\meter}, $\alpha(\lambda_2) = 7.2 \times 10^{-5}$\,\si{\dB/\meter}), each with $\approx 40\,\si{\nano\meter}$ FWHM bandwidth. 
The depth maps neglecting and including the air emission are depicted in Fig.~\ref{fig:Bispectral_Ranging2}\subref{fig:Air_emission_Neglecting_air} and Fig.~\ref{fig:Bispectral_Ranging2}\subref{fig:Air_emission_Including_air}, respectively.
Fig.~\ref{fig:Bispectral_Ranging2}\subref{fig:Air_emission_lidar} shows the depth map of the scene from lidar for comparison, although pixel-to-pixel correspondence between the sensors is only approximate. 
Fig.~\ref{fig:Bispectral_Ranging2}\subref{fig:Air_emission_Temperature_difference} shows the estimated temperature difference with respect to air.
For estimating the temperature difference map we use the brightness temperature at the clear band (8.38\,\si{\micro\meter}), assuming the emissivity is 1\@.
We use the weather station data collected on-site for the air temperature.
Based on this analysis, most of the objects in the scene appear to be about the same temperature or cooler than the air.
Looking at Fig.~\ref{fig:Bispectral_Ranging2}\subref{fig:Air_emission_Neglecting_air}, neglecting the air emission results in very poor estimates.
In most of the first image, the colors range from magenta to white, as the estimator returns negative or close-to-zero range estimates for most of the scene.
The results in Fig.~\ref{fig:Bispectral_Ranging2}\subref{fig:Air_emission_Including_air}, where air emission is included in the model, shows ranging results much closer to those in the lidar depth map shown in Fig.~\ref{fig:Bispectral_Ranging2}\subref{fig:Air_emission_lidar}.

Fig.~\ref{fig:Bispectral_ranging_time}(a) shows temperature difference and depth map reconstructions of the same scene collected over 90 minutes in 10 minute time intervals. 
Fig.~\ref{fig:Bispectral_ranging_time}(b) shows the temperature analysis of the scene throughout the 90 minutes.
The object temperatures in the scene are plotted with gray diamonds, pseudorandomly spread horizontally for visualization purposes.
A histogram of the scene temperature for the first acquisition is also plotted on the right.
There are two main thermal regions in the scene, the foreground rolling grass area, highlighted with red and the background forest area highlighted with blue.
The rolling grassy area is the coolest throughout the time of the data collection with around -4\,\si{K} relative temperature with respect to air.
The background forest is near air temperature and there are parts that are cooler and hotter than air.
Ideally, the air temperature should be estimated from a saturated ($\tau \approx 0$) absorptive band; however, the 8--13\,\si{\micro\meter} LWIR band is not absorptive enough to estimate the air temperature.
Instead, weather station readings are used as initial estimates for air temperature, plotted with red circles at the data collection times.
We then manually adjust the model air temperature for each data set to achieve good qualitative agreement with the lidar data. 
We adjust the model air temperature no more than 1\,\si{K} from the weather station temperature to get the results in Fig.~\ref{fig:Bispectral_ranging_time}(a), shown with yellow circles.
Considering the bispectral reconstruction on the bottom row, the range gradient in the grassy area from bottom to top appears consistently in all estimates throughout the 90 minute data collect, even as grass and air temperatures change.
The background forest area, where the temperature difference is low, is not as consistent.
This is consistent with the Fisher information analysis in Section~\ref{sec:FisherInformationAnalysis},
which showed the distance estimation problem to be more challenging when
temperature difference is low.

\begin{figure}
    \centering
    \includegraphics[width=.5\textwidth]{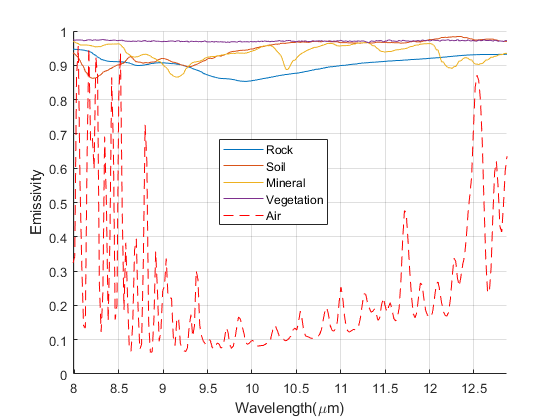}
    \caption{Emissivity profiles of solid objects from the ECOSTRESS database~\cite{MEERDINK2019111196, BALDRIDGE2009711} and of the atmosphere along 1000\,\si{\meter} distance with 289.7\,\si{K} temperature, 1010\,\si{\milli bar} pressure, and 0.012 VMR water vapor from HITRAN~\cite{GORDON2022107949}.
    The emissivity profiles of solid objects are smooth and continuous whereas the emissivity of the atmosphere has sharp features.}
    \label{fig:EmissivityReg}
\end{figure}

\begin{figure*}
    \centering
    \subfloat[Rock\\ $\dhatReg = 100.03\,\si{\meter}$ \\ $\dhatNoReg = 5.77\,\si{\meter}$]{
        \centering
        \includegraphics[width=0.33\textwidth]{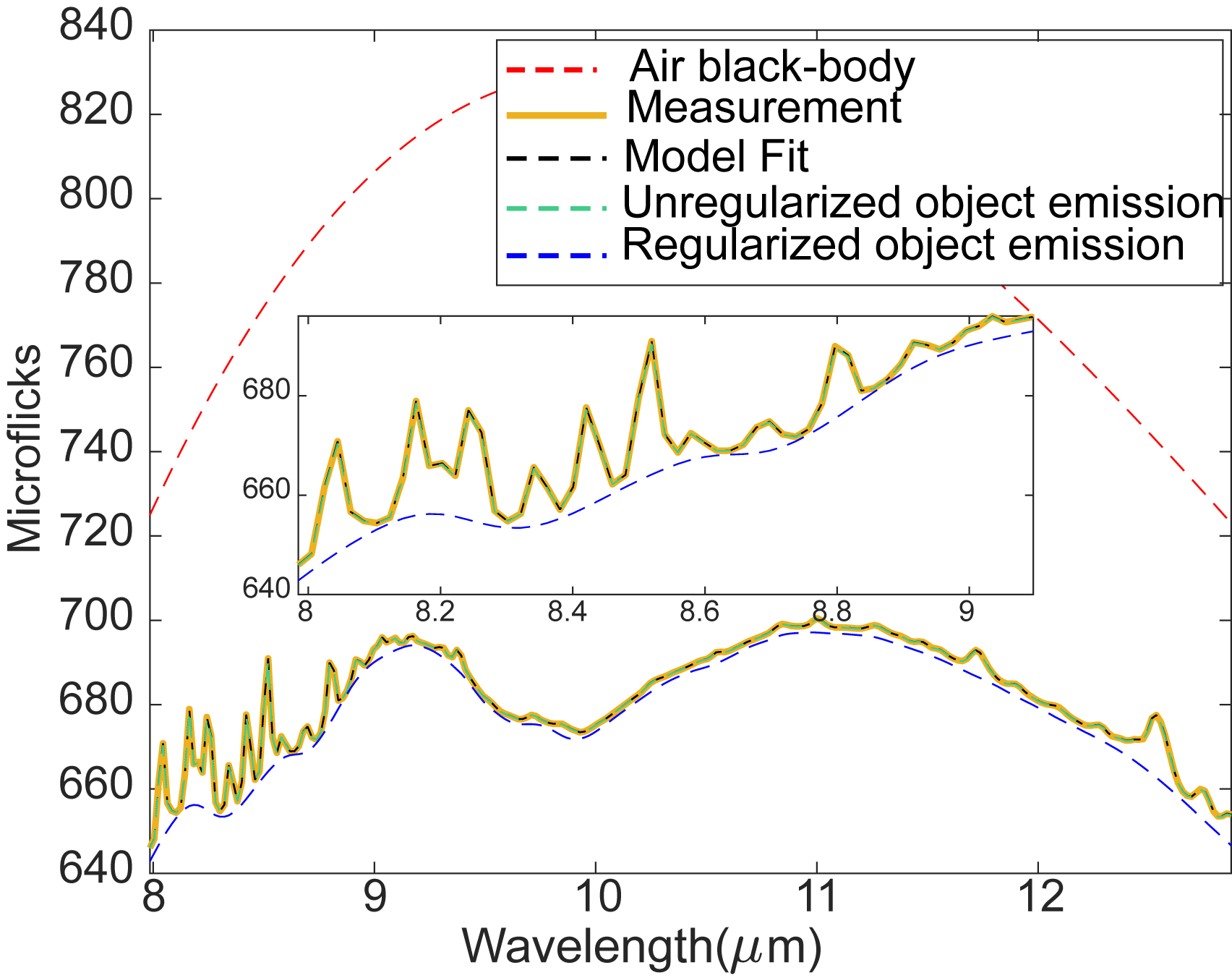}
        \label{fig:Rockd100NoNoiseReg}
    }
    \subfloat[Soil\\ $\dhatReg = 99.53\,\si{\meter}$ \\ $\dhatNoReg = 9.27\,\si{\meter}$]{
        \centering
        \includegraphics[width=0.33\textwidth]{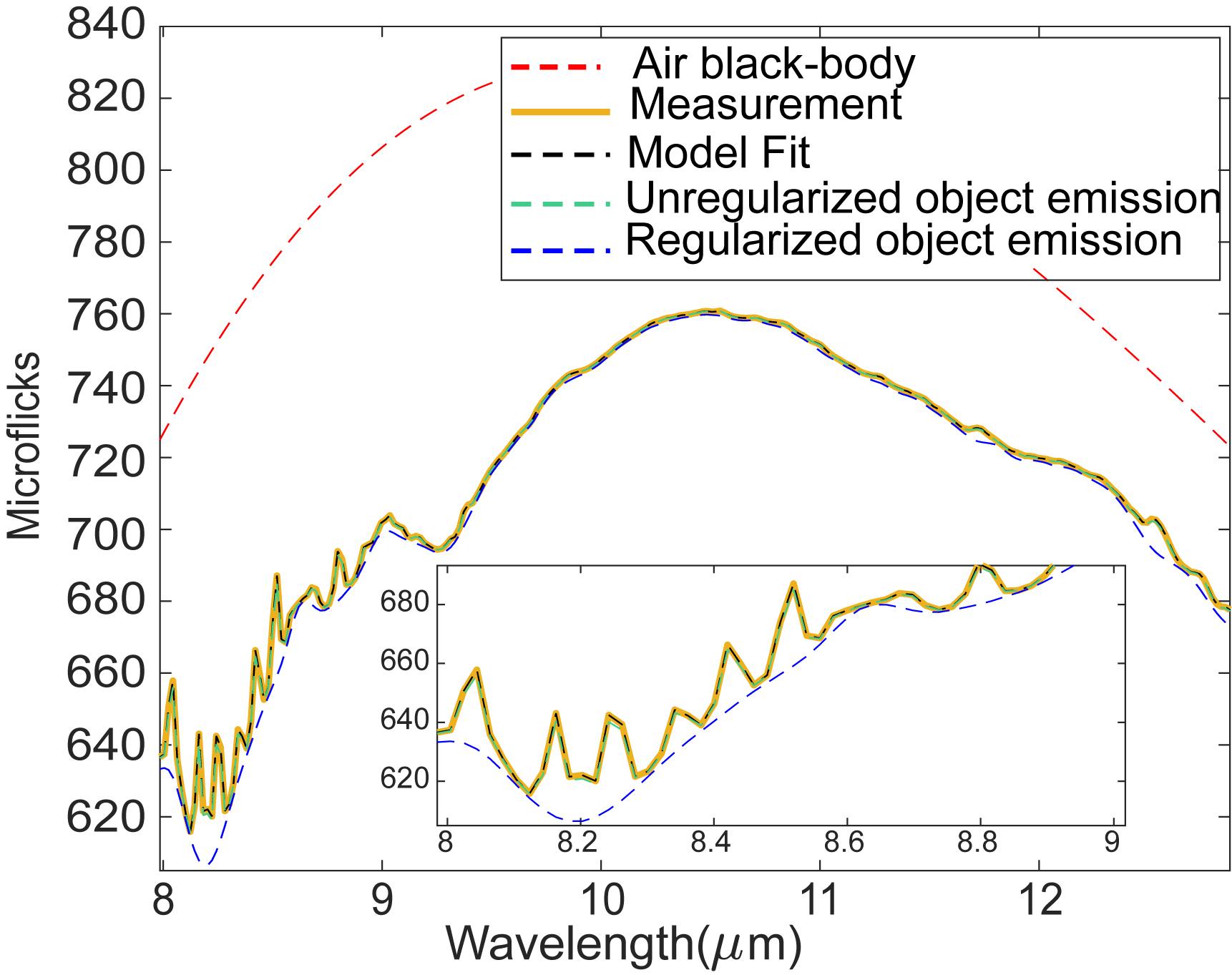}
        \label{fig:Soild100NoNoiseReg}
    }
    \subfloat[Vegetation\\ $\dhatReg = 100.03\,\si{\meter}$ \\ $\dhatNoReg = 37.57\,\si{\meter}$]{
        \centering
        \includegraphics[width=0.33\textwidth]{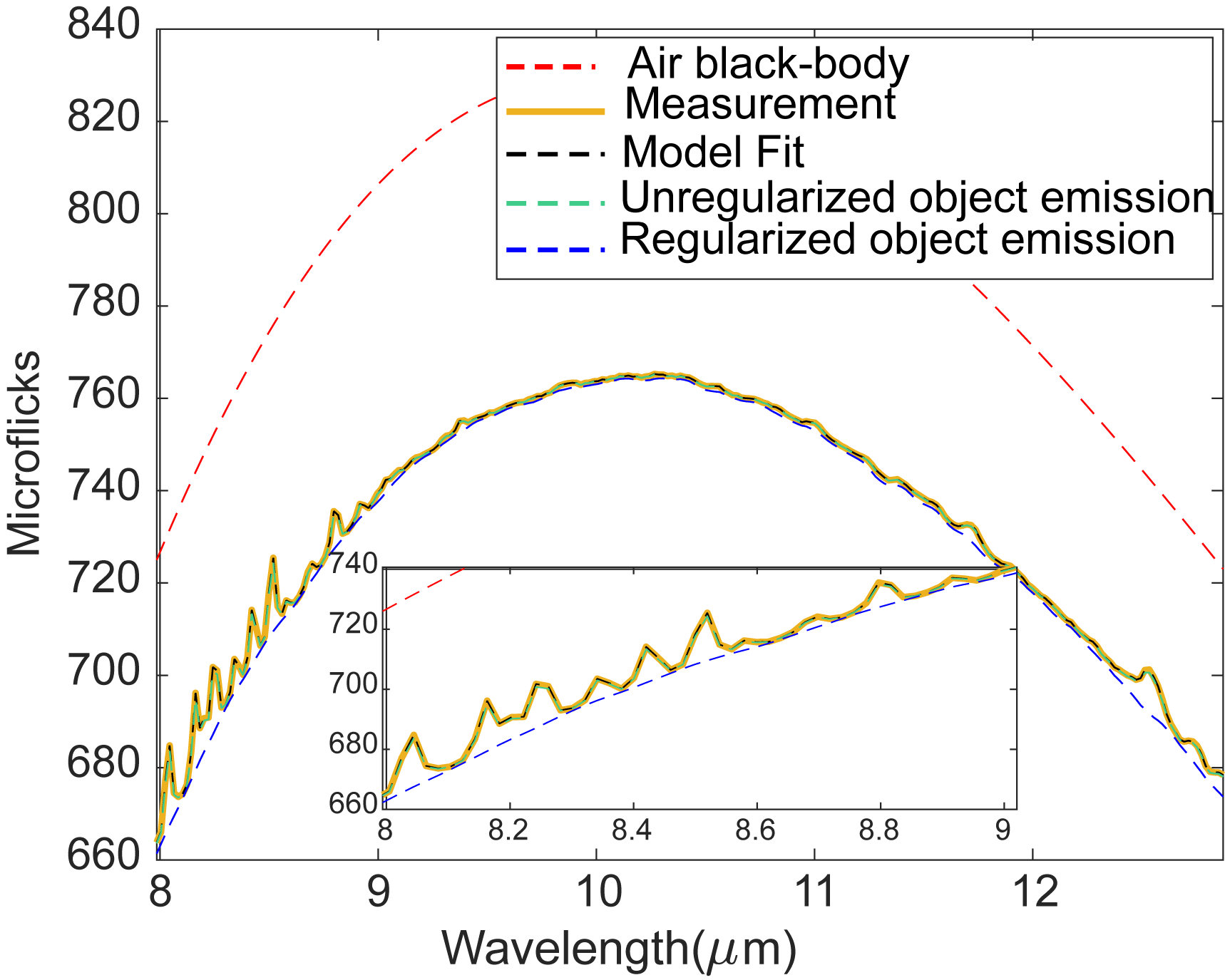}
        \label{fig:Vegetationd100NoNoiseReg}
    }
    \\
    \subfloat[Rock]{
        \centering
        \includegraphics[width=0.32\textwidth]{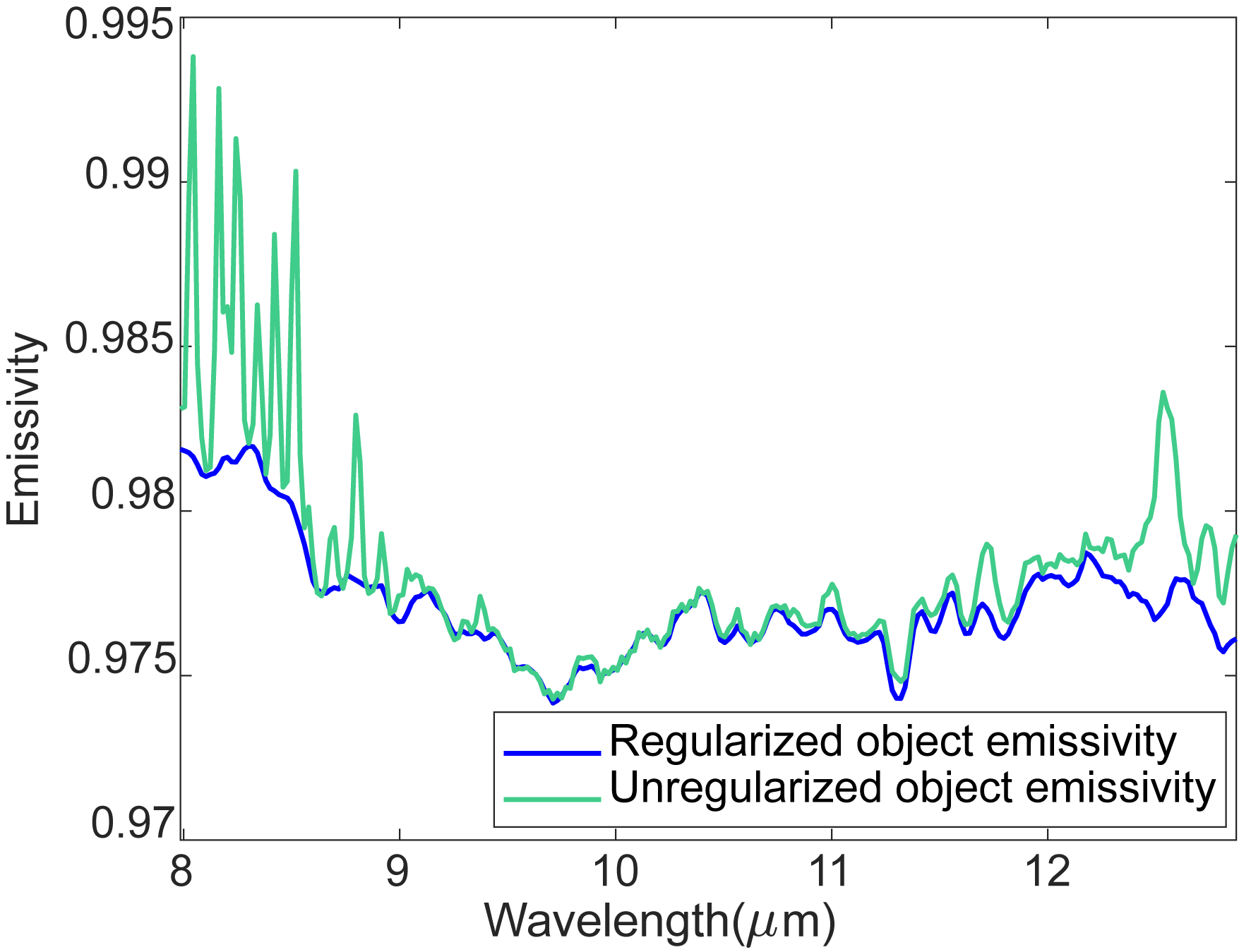}
        \label{fig:Rockd100NoNoiseEmissivity}
    }
    \subfloat[Soil]{
        \centering
        \includegraphics[width=0.32\textwidth]{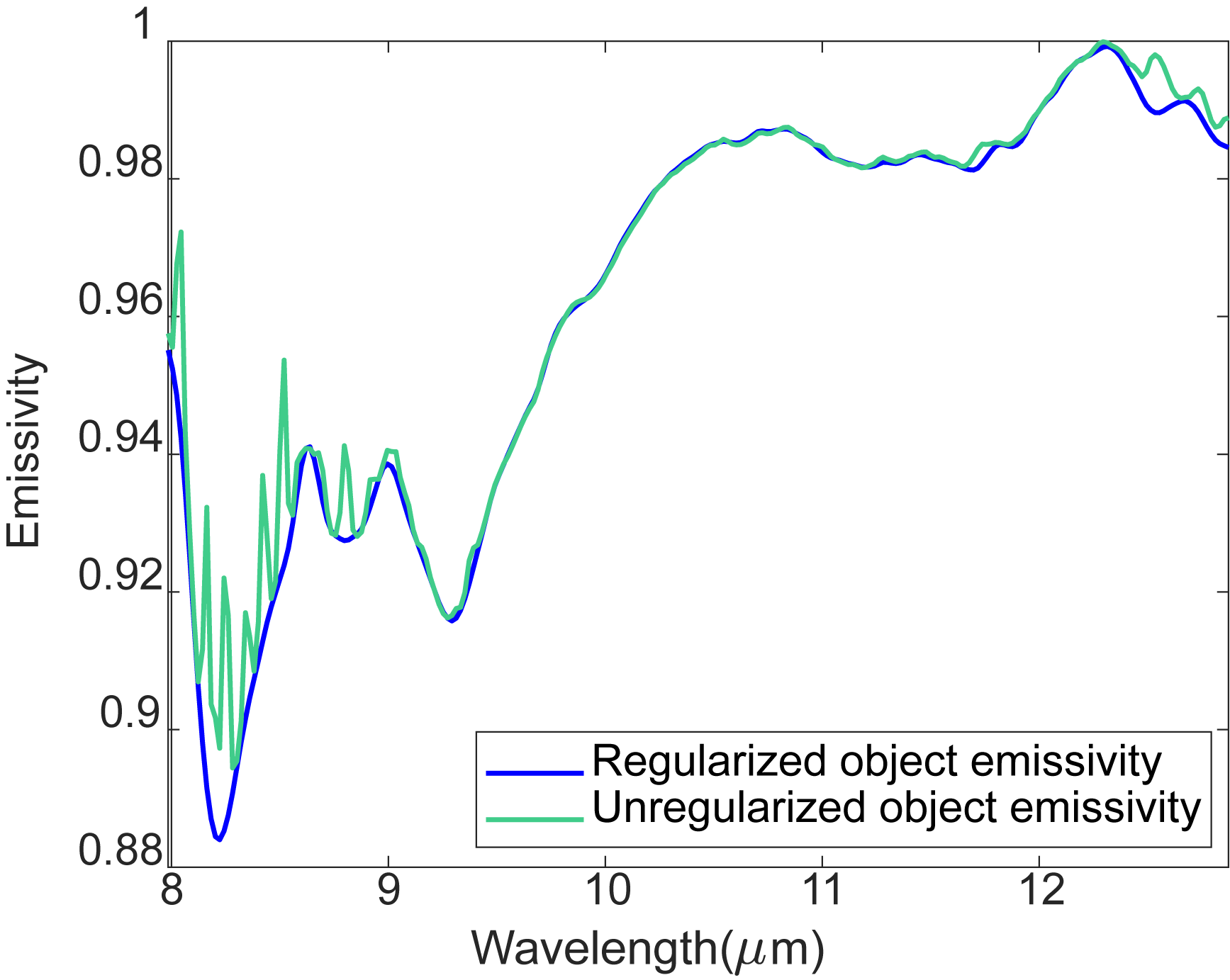}
        \label{fig:Soild100NoNoiseEmissivity}
    }
    \subfloat[Vegetation]{
        \centering
        \includegraphics[width=0.33\textwidth]{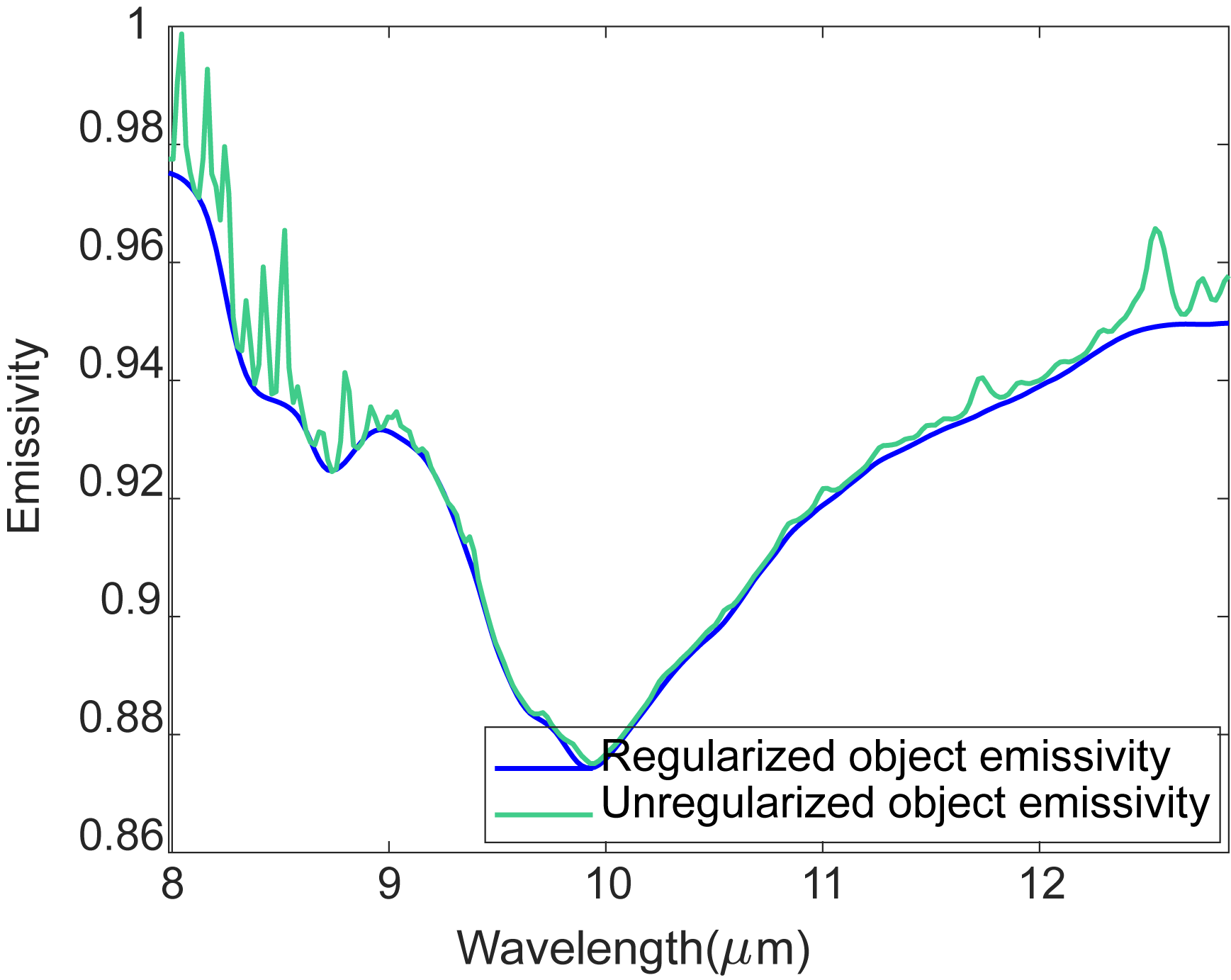}
        \label{fig:Vegetationd100NoNoiseEmissivity}
    }
    \caption{Simulation results for noiseless hyperspectral ranging for three example objects, each 100\,\si{\meter} away from the sensor.
    The air is at 289.7\,\si{K} temperature, whereas each object is 3\,\si{K} cooler ($\Delta T = -3$\,\si{K}).
    The object emissivity profiles are chosen from three categories of the ECOSTRESS~\cite{MEERDINK2019111196, BALDRIDGE2009711} database: rock, soil, and vegetation.
    \protect\subref{fig:Rockd100NoNoiseReg}, \protect\subref{fig:Soild100NoNoiseReg}, and \protect\subref{fig:Vegetationd100NoNoiseReg} compare estimates computed with and without the emissivity regularization in \eqref{eq:Lreg}; 
    in each case, the measurement, the model fit, and the unregularized object emission estimate are almost identical to one another.
    The distance estimates without regularization are very poor.
    \protect\subref{fig:Rockd100NoNoiseEmissivity}, \protect\subref{fig:Soild100NoNoiseEmissivity}, and \protect\subref{fig:Vegetationd100NoNoiseEmissivity} show the emissivity estimates with (blue) and without (cyan) regularization.
    The regularized estimator finds reasonable solutions,
    whereas the unregularized estimator does not because it ascribes much of the absorption structure to the object emissivity estimate.}
    \label{fig:SimulationResultsNoiseless}
\end{figure*}


\section{Hyperspectral Ranging}
\label{sec:hyperspectral}
This section describes an absorption-based range imaging method using the full LWIR spectral information.
Joint estimation of distance, emissivity, and temperature is important when processing a wide spectral window.
In contrast to the assumption in \eqref{eq:SameEmissivity}, we cannot assume constant object emission, as emissivity profiles are wavelength dependent.  

Inversion of the forward model in \eqref{eq:ForwardModelTransmittance2} is underdetermined. 
For measurements at $K$ spectral channels, there are $2K+2$ unknowns ($K$ unknowns from transmittance, $K$ unknowns from emissivity, and $2$ unknowns from air and object temperature). 
To solve the inverse problem, we introduce additional assumptions. 
First, the air temperature is assumed to be known, from either a highly absorptive band or another sensor.
Second, assuming air pressure and water vapor content are also known, the attenuation function $\alpha(\lambda)$ is then fixed for all $\lambda$. Only the distance $d$ in \eqref{eq:Trasnmittance} remains to be determined, reducing the number of unknowns in the transmittance function from $K$ to $1$.
The use of a known attenuation function exploits the spectral structure of transmittance.
With these assumptions, for $K$ spectral measurements per scene pixel, there are $K+2$ unknowns
($K$ unknowns from the object emissivity, $1$ unknown from the object temperature and $1$ unknown from the object distance).
The problem is still underdetermined, and an emissivity-smoothing regularizer is used to promote a good solution.

The inversion is formulated as an optimization problem, where the loss function to be minimized has a data fidelity term and a regularizer term. 

The data fidelity term is the squared $L_2$ norm between the model prediction and the measurement:
\begin{equation}
    \Ldata(d,T,\bm{\emiss}) = \sum_{k = 1}^K (\yhat_{k}(d,T,\emiss_k) - y_{k})^2, 
    \label{eq:LData}
\end{equation}
where $d \in \mathbb{R}$, $\bm{\emiss} \in \mathbb{R}^K$, $T \in \mathbb{R}$, and $\yhat_k(d,T,\emiss_k)$
is computed using the forward model in \eqref{eq:ForwardModel}.

A regularization term is used to incorporate our knowledge about typical object emissivity profiles. 
Fig.~\ref{fig:EmissivityReg} shows the emissivity profiles of the surface atmosphere and examples of some solid objects from several categories including rock (diorite gneiss), soil (gravelly loam), mineral (rivadavite), and vegetation (abies concolor).
The emissivity profiles of solid objects are smooth compared to the sharp transitions in the atmospheric transmittance \cite{8738014, borel2008error, borel2003artemiss}.
This knowledge is incorporated into the loss function by using a Tikhonov regularization with a difference matrix $D$:
\begin{equation}
\label{eq:Lreg}
    \Lreg(\bm{\emiss})
        = \|D\bm{\emiss}\|_2^2
        = \sum_{k = 1}^{K-1} (\emiss_{k+1} - \emiss_k)^2.
\end{equation}
Our final loss function combines terms from data fidelity and regularization:
\begin{align}
  \mathcal{L}(&d,T,\bm{\emiss})
     = \Ldata(d,T,\bm{\emiss})  +  \rho \Lreg(\bm{\emiss}) \nonumber \\
    &= \sum_{k = 1}^K (\yhat_{k}(d,T,\emiss_k) - y_{k})^2 + \rho \sum_{k = 1}^{K-1} (\emiss_{k+1} - \emiss_k)^2 ,
\end{align}
where $\rho$ controls the strength of the regularization.

Our optimization is performed for each scene pixel independently, incorporating spectral regularization.
Spatial regularization---such as adding total variation (TV) penalties on temperature, emissivity or distance---could be incorporated for denoising purposes, including horizontal stripe artifacts in the measurements, at the cost of more memory and computational requirements.

A closed-form solution to minimize the loss function was not found.
Instead, an iterative gradient descent method is used to estimate the parameters from the optimization problem.
First, the attenuation function $\alpha(\lambda)$ is calculated from weather station readings of air temperature, humidity, and pressure.
Then, the regularized optimization problem is solved for each pixel separately.
For each pixel, an emissivity profile, a temperature value, and a distance value are estimated.

The implementation processes $64 \times 64 \times 256$ (Height $\times$ Width $\times$ Spectral Channels) patches in parallel. The memory usage is approximately $40\,\si{\mega\byte}$, and it takes around $1\,\si{\minute}$ to optimize a single patch. The bispectral method uses two spectral channels and has a closed-form solution, making it significantly faster. For an input size of $260 \times 1600 \times 2$ (Height $\times$ Width $\times$ Spectral Channels), it takes approximately $70\,\si{\milli\second}$ and requires around $16\,\si{\mega\byte}$ of memory.
All timings were measured on a system with an Intel Core i9-12900K CPU, a GeForce RTX 3060 GPU with 12GB GDDR6 memory, and 64GB DDR5 RAM.

\begin{figure*}
        \centering
        \subfloat[$\Delta T = -8$\,\si{K}]{\includegraphics[width=.33\textwidth]{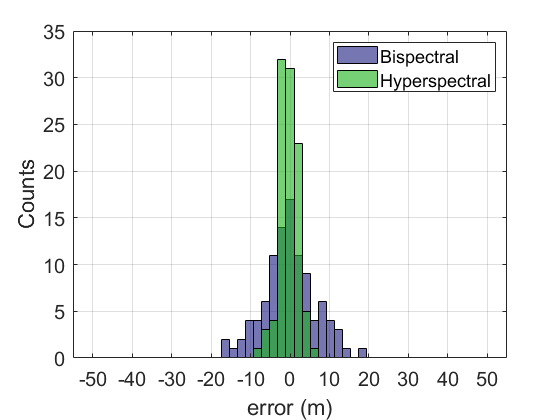}\label{fig:T8S1}} \hfill
        \subfloat[$\Delta T = -5$\,\si{K}]{\includegraphics[width=.33\textwidth]{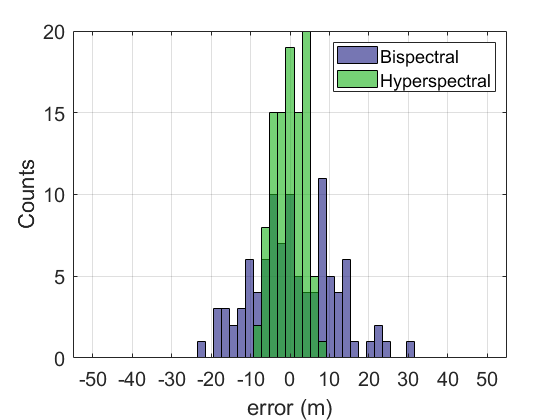}\label{fig:T5S1}} \hfill
        \subfloat[$\Delta T = -2$\,\si{K}]{\includegraphics[width=.33\textwidth]{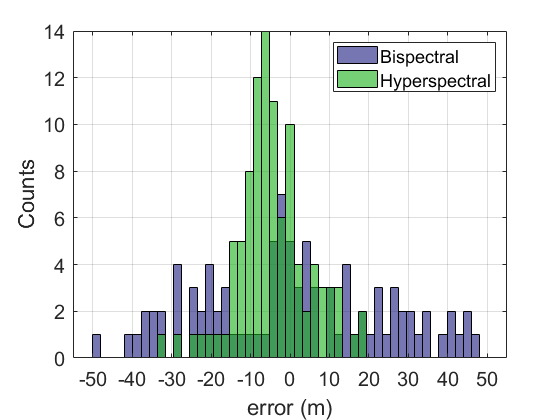}\label{fig:T2S1}}
        \caption{Simulation results of bispectral (purple) and hyperspectral (green) ranging methods under noisy observations with additive white Gaussian noise of standard deviation of 1 microflick ($\sigma = 1$).
        The effect of noise increases with decreasing absolute temperature difference.
        Numerical comparison is in Table \ref{tab:Simulations_table}.
        }
        \label{fig:NoisyMeasurementsAnalysis}
\end{figure*}

\section{Results}
\label{sec:results}
The method of Section~\ref{sec:hyperspectral} was tested both on simulated and experimental data.
Simulations were conducted in both noiseless and noisy scenarios.
The spectral regularization is needed to find the correct solution.
For noiseless simulated data, the regularized estimator can extract object emissivity and range with almost no errors.
For noisy simulated data, the hyperspectral estimator is analyzed for a variety of simulated temperature differences.
Finally, we validate our method with experimental data and analyze the ranging performance.

\subsection{Simulation Results}
\label{sec:simulation}
We first present simulation results with noiseless data to highlight the necessity of the regularization term introduced in \eqref{eq:Lreg}.
The forward model in \eqref{eq:ForwardModel} is used to simulate spectral measurements for an object at a distance of 100\,\si{\meter}.
The atmospheric parameters and spectral resolution are matched to the conditions of the scene shown in Fig.~\ref{fig:ExperimentalResultsP5S1} with 289.7\,\si{K} air temperature, 1010\,\si{\milli bar} air pressure, 0.012 VMR water vapor, and 40\,\si{\nano\meter} Gaussian ISRF.
The object is assumed to be 3\,\si{K} cooler than the air.
We use emissivity profiles from the ECOSTRESS database~\cite{MEERDINK2019111196, BALDRIDGE2009711}.

The distance is estimated with and without regularization using a gradient descent initialized at constant values,
\begin{align*}
    \That^{(0)}       &= 300\,\si{K}, \\
    \emisshat^{(0)}_k &= 0.9, \\
    \dhat^{(0)}       &= 150\,\si{\meter}.
    \label{eq:Initialization}
\end{align*}
The regularization parameter $\rho$ is set to $10^{7}$ for the regularized case, and we stop after $20000$ iterations.

Fig.~\ref{fig:SimulationResultsNoiseless} shows results for example emissivity profiles from three categories: rock (diorite gneiss), soil (gravelly loam), and vegetation (abies concolor).
The first row shows the measurements and the model fits.
In each of these plots,
the red dashed curve represents a black-body at air temperature,
the yellow curve represents the simulated measurement,
and the black dashed curve represents the fit to the measurement.
Only one black dashed curve is provided because
the fits for the regularized and unregularized cases are nearly identical.
The estimated object emissions are represented with blue dashed curves for regularized and cyan dashed curves for unregularized cases.
Notice that the regularized estimator gives almost exactly the correct distance of 100\,\si{\meter} in all three examples, whereas the unregularized estimator gives large errors.
The unregularized object emission estimate is almost identical to both the measurements and the model fit.
This is because, without regularization, most of the sharp spectral features in the measurements are incorrectly attributed to the object emission term, rather than to atmospheric absorption.

The second row provides plots of estimated emissivities to more clearly illustrate the difference between the regularized and unregularized estimators.
Without regularization, the solution is not unique.
For many estimated distances, the estimator can ascribe all spectral features remaining after factoring out transmission effects to the emissivity.
The regularized estimator finds reasonable solutions for each of the three emissivity profiles.
Errors in the distance estimation would propagate into the estimated emissivity profile in the form of sharp features.
The regularization disfavors these sharp features in the emissivity, thus encouraging accurate distance estimates.

Simulations were also conducted assuming noisy measurements.
Fig.~\ref{fig:NoisyMeasurementsAnalysis} shows the performance under additive white Gaussian noise (AWGN)
with standard deviation of 1 microflick for a rock 100\,\si{\meter} away from the sensor. 
We keep the atmospheric parameters fixed.
We repeat our Monte Carlo experiments 100 times so that we can find empirical distributions of the distance error.
Moving from left to right, the relative temperature is increasing from $-8$\,\si{K} to $-5$\,\si{K} to $-2$\,\si{K}.
As the temperature difference decreases, the effect of noise increases.
This is because the spectral measurements are more sensitive to range when the temperature difference is large, as was explained in Section~\ref{sec:FisherInformationAnalysis}.
Fig.~\ref{fig:NoisyMeasurementsAnalysis} also compares the performance of the bispectral and hyperspectral methods.
As expected, the bispectral method performs worse than the hyperspectral method.
From Fig.~\ref{fig:NoisyMeasurementsAnalysis}\subref{fig:T2S1}, it is clear that when the temperature difference is low, it is crucial to use all of the available information.
Table~\ref{tab:Simulations_table} shows the comparison of root-mean-squared error (RMSE) for these two methods. 
The hyperspectral method shows an RMSE 2.5 times smaller than the bispectral method.

\begin{table}
\caption{Root mean-squared error comparison of bispectral and hyperspectral methods on noisy measurements of a rock at 100\,\si{\meter};
histograms shown in Fig.~\ref{fig:NoisyMeasurementsAnalysis}.
}
\label{tab:Simulations_table}

\centering
\scalebox{0.92}{
\begin{tabular}{rrrr}
\toprule
$\Delta T$    &              -8\,\si{K} &              -5\,\si{K} &         -2\,\si{K}\\[0.1ex]
\midrule
Bispectral    &      $6.5$\,\si{\meter} &     $10.4$\,\si{\meter} &      $27.2$\,\si{\meter} \\
Hyperspectral & $\bm{2.4}$\,\si{\meter} & $\bm{3.8}$\,\si{\meter} & $\bm{10.1}$\,\si{\meter} \\
\bottomrule
\end{tabular}
}
\end{table}

\subsection{Experimental Results}
\label{sec:experimental}

\subsubsection{Data Specifications}
\label{sec:experimental_data_spec}
The experimental data provided by the U.S. Army Night Vision and Electronic Sensors Directorate and the Johns Hopkins University Applied Physics Laboratory were acquired using a pushbroom LWIR hyperspectral imager with a cooled HgCdTe sensor.
The pushbroom sensor collects the datacube using a 2D sensor array that spans the spectrum and vertical axis. The sensor then scans horizontally across the scene to acquire the full datacube. This horizontal scanning strategy can result in stripe artifacts in the collected data due to non-uniform pixel responses or malfunctioning pixels in the spectral domain. These artifacts are visible in the animation included in the supplementary material, which presents the 2D slices across the wavelength dimension.
The spectrometer has a vertical field-of-view (VFOV) of 11.6 degrees and a horizontal FOV (HFOV) of 57 degrees.
The focal length is 50 mm and the f-number is f/0.9\@.
The typical noise of the sensor is around 1 microflick, which---at 10\,\si{\micro \meter} wavelength---corresponds to about a 1000:1 signal-to-noise ratio~\cite{bao2023heat}.
For each of the $1280 \times 260$ pixels in the scene, the sensor acquires data for 256 spectral bands between 8.0\,\si{\micro\meter} and 13.2\,\si{\micro\meter}\@. The spectral resolution is $\approx 40\,\si{\nano\meter}$.
Before analysis, the raw data underwent a wavelength correction of $-120\si{\nano\meter}$ to address a calibration mismatch in the sensing instrument, aligning the measurements more closely with known atmospheric absorption features.

During the data collection, an on-site weather station also recorded humidity, air temperature and pressure levels.
These three parameters were used to calculate the attenuation function with high-resolution spectral modeling software (Spectralcalc)~\cite{SpectralCalc}.
We assume that these parameters are constant over the scene and use a single attenuation function and air temperature for all pixels.
Lidar measurements were also acquired using a high-resolution system that collects 1240 points over 360$^{\circ}$, with a point spacing of $\sim$6\,\si{\milli\meter} at 10\,\si{\meter}, a VFOV of 150$^{\circ}$ and an HFOV of $\approx 90^{\circ}$;
the system has ranging uncertainty of $\pm$1\,\si{\milli\meter} and a maximum range of 350\,\si{\meter}. 

The experimental data were collected
from 19:41 to 21:01 local time, with 10 minutes of separation between data collects;
sunset was at 19:44\@.
The scene mostly consists of rolling grassy terrain.
There are sky and forest pixels at the top of the VFOV\@.
There is a tree in the right foreground of the scene.
There are also 7 calibration targets with checkerboard patterns located throughout the scene.

\begin{figure*}
    \centering
    \subfloat[Absorption-based ranging]{\includegraphics[width=.85\textwidth]{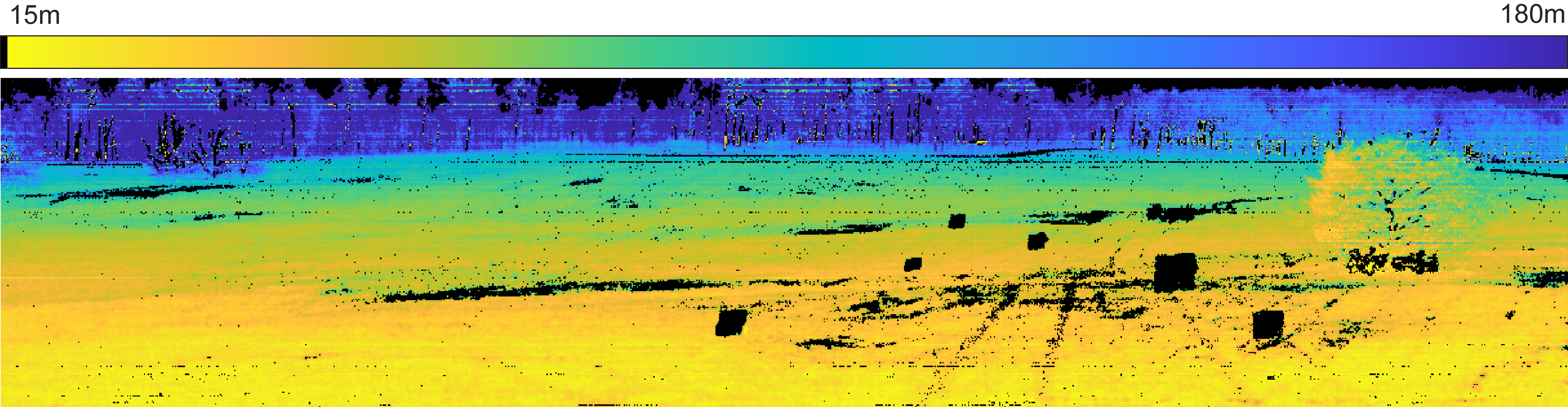}\label{fig:Experimental_results_Depth_map}} \hfill
    \subfloat[Lidar]{\includegraphics[width=.85\textwidth]{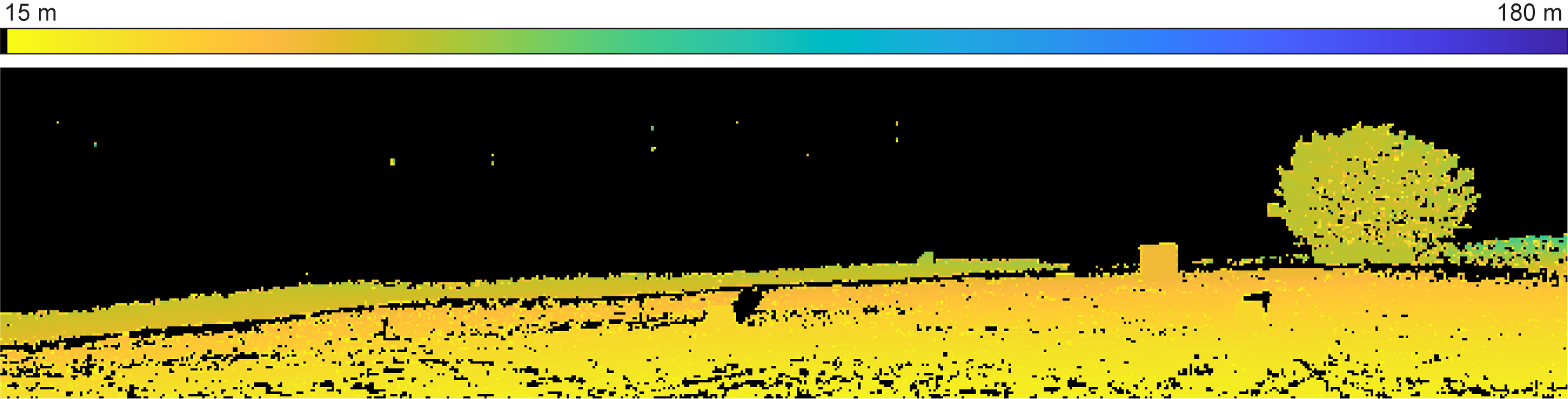}\label{fig:Experimental_results_Lidar}} \hfill
    \subfloat[Emissivity clusters]{\includegraphics[width=.85\textwidth]{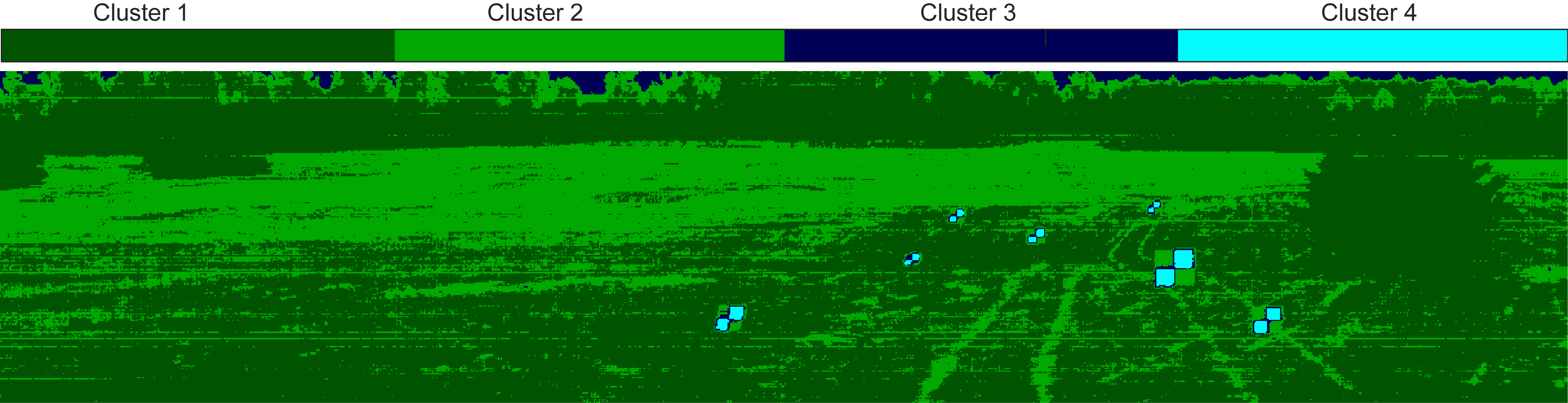}\label{fig:ExperimentalResultsEmissivityClusters}}\hfill
    \subfloat[Temperature map]{\includegraphics[width=.85\textwidth]{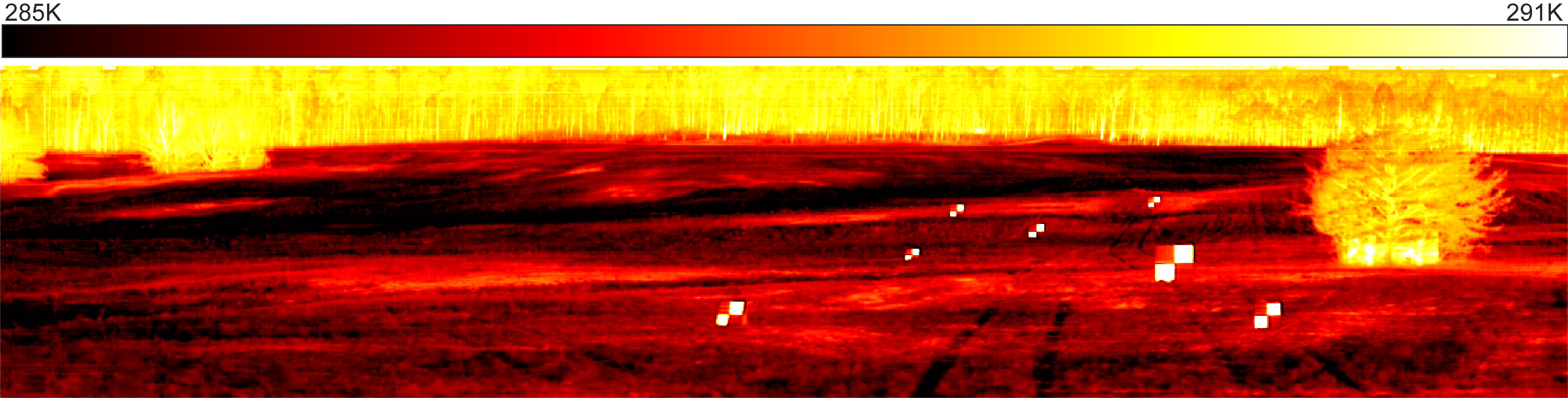}\label{fig:Experimental_results_Temperature_map}}\hfill
    \subfloat[RGB image]{\includegraphics[width=.85\textwidth]{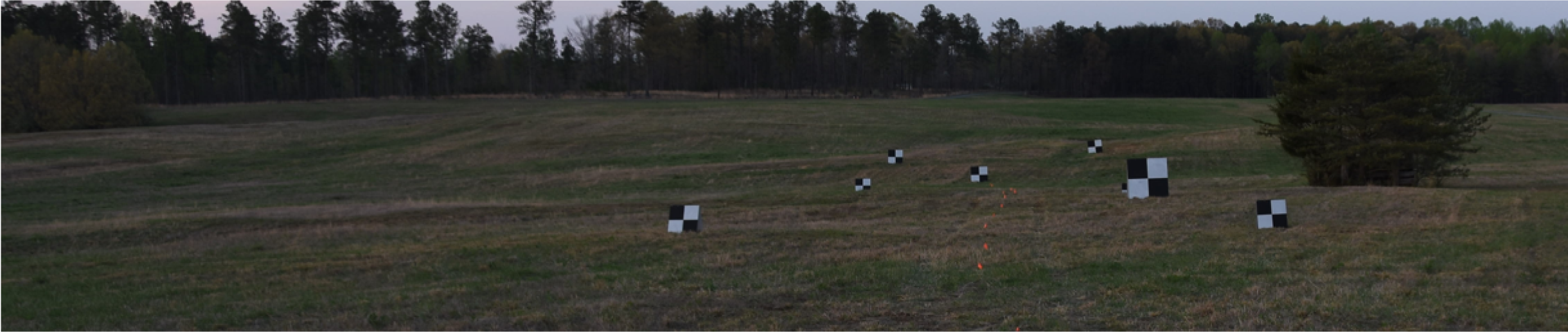}\label{fig:Experimental_results_RGB_image}}
    \caption{Results on experimental data using hyperspectral ranging: \protect\subref{fig:Experimental_results_Depth_map} depth map; \protect\subref{fig:Experimental_results_Lidar} depth map acquired via lidar for comparison, although pixel-to-pixel correspondence is only approximate; \protect\subref{fig:ExperimentalResultsEmissivityClusters} a $k$-means clustering of the emissivity estimates; \protect\subref{fig:Experimental_results_Temperature_map} an estimated temperature map; and \protect\subref{fig:Experimental_results_RGB_image} an RGB image of the scene for visual reference.
    }
    \label{fig:ExperimentalResultsP5S1}
\end{figure*}

\begin{figure}
    \centering
    \includegraphics[width=.45\textwidth]{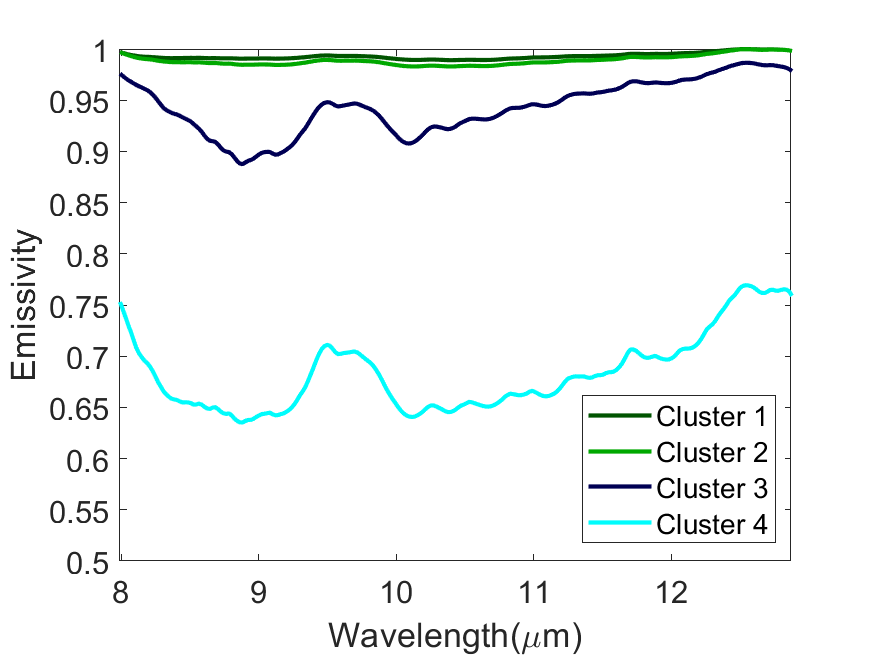}
    \caption{Mean emissivity profiles from $k$-means algorithm.
    The colors of the curves match with Fig.~\ref{fig:ExperimentalResultsP5S1}\protect\subref{fig:ExperimentalResultsEmissivityClusters}.
    Clusters~1 and~2 are predominantly vegetation, close to black-body.
    Cluster~3 is mostly the background sky, and Cluster~4 is largely the reflective panels of the calibration targets.
    The emissivity profiles for these show large ozone absorption feature at 9.6\,\si{\micro\meter}, presumably due to the reflected downwelling radiance component.
    We associate the overestimations with significant downwelling contributions.}
    \label{fig:EmissivityClusters}
\end{figure}

\subsubsection{Hyperspectral Ranging Results}
\label{sec:experimental_hyperspectral_results}
The results of the hyperspectral method on experimental data are summarized in Fig.~\ref{fig:ExperimentalResultsP5S1}.
The extracted depth map is in Fig.~\ref{fig:ExperimentalResultsP5S1}\subref{fig:Experimental_results_Depth_map}, where black pixels represent unreliable points discussed later.
Fig.~\ref{fig:ExperimentalResultsP5S1}\subref{fig:Experimental_results_Lidar} shows a depth map acquired with lidar that can be used to validate our results; black pixels represent points with no data.
Overall, our method matches a great portion of the scene.
In addition to the distance estimation, we also estimate the emissivity profile for each pixel.
Fig.~\ref{fig:ExperimentalResultsP5S1}\subref{fig:ExperimentalResultsEmissivityClusters} shows the results of clustering the estimated emissivity profiles with $k$-means of 4 clusters. 
Fig.~\ref{fig:EmissivityClusters} shows the emissivity profiles for each cluster mean.
Objects of similar type are largely clustered together.
Clusters 1 and 2 are predominantly vegetation, close to black-body.
Cluster 3 is mostly background sky.
Cluster 4 is largely the reflective panels of the calibration targets.
The estimated object temperatures are shown in Fig.~\ref{fig:ExperimentalResultsP5S1}\subref{fig:Experimental_results_Temperature_map}, and Fig.~\ref{fig:ExperimentalResultsP5S1}\subref{fig:Experimental_results_RGB_image} shows the RGB image of the scene for reference.

\begin{figure}
    \centering
    \subfloat[Difference image around ozone absorption at 9.6\,\si{\micro\meter}
    ]{\includegraphics[width=.5\textwidth]{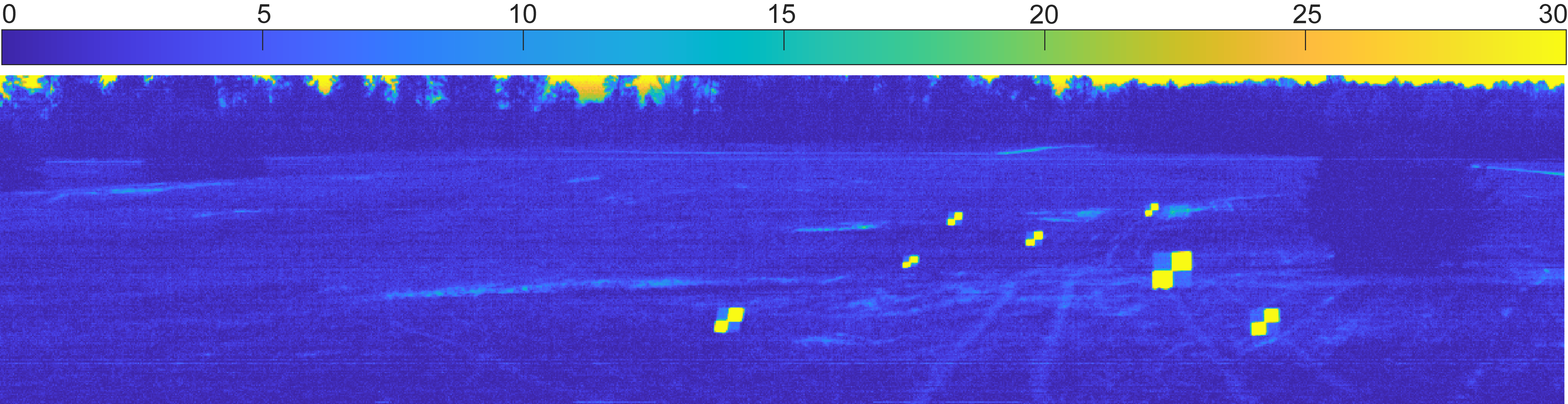}\label{fig:Ozone-01}} \hfill
    \subfloat[Binary mask based on thresholding the ozone difference image]{\includegraphics[width=.5\textwidth]{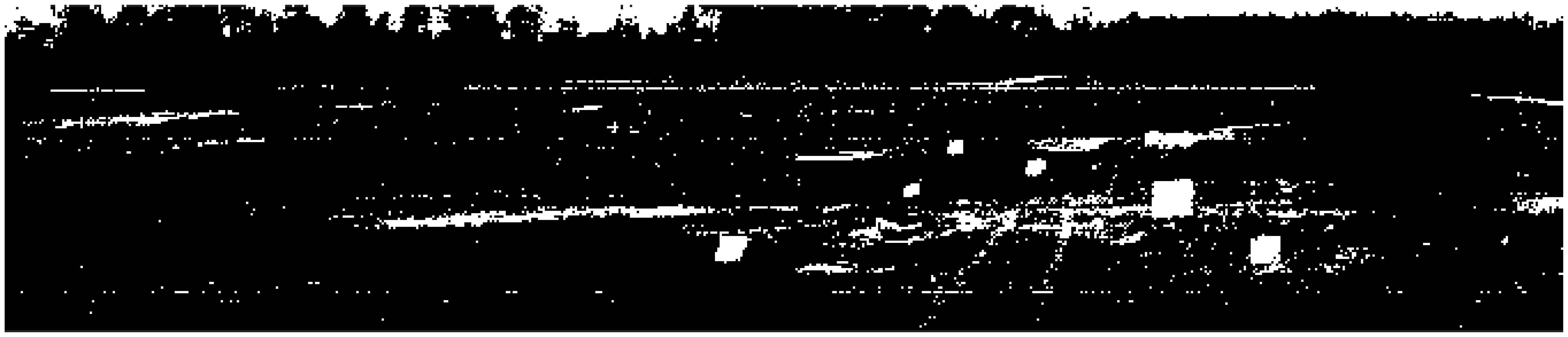}\label{fig:Ozone-02}} \hfill
    \subfloat[Range estimate for all pixels
    ]{\includegraphics[width=.5\textwidth]{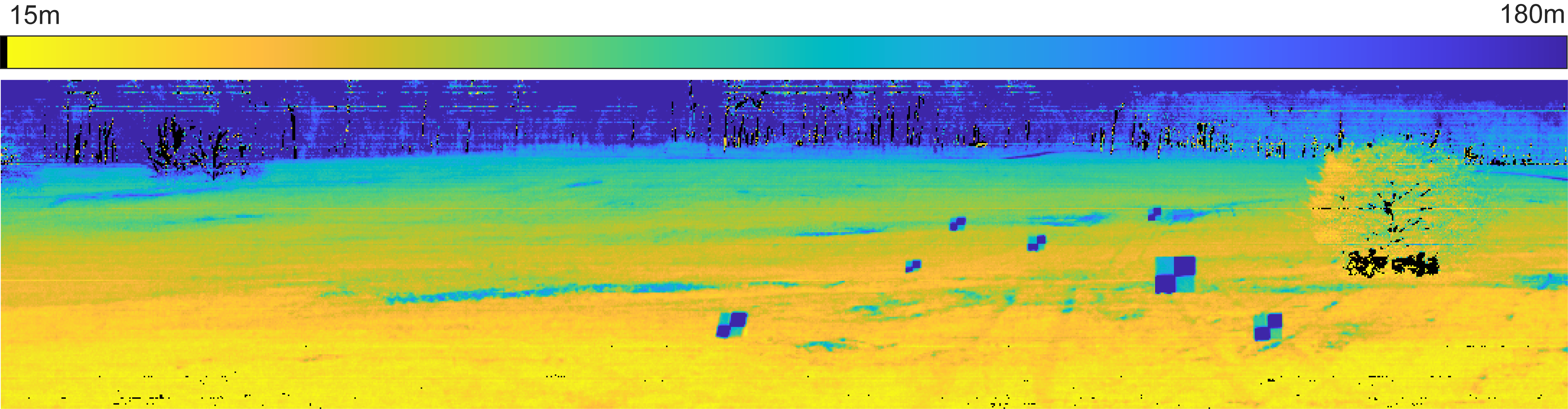}\label{fig:Ozone-03}} \hfill
    \subfloat[Range estimate for reliable pixels
    ]{\includegraphics[width=.5\textwidth]{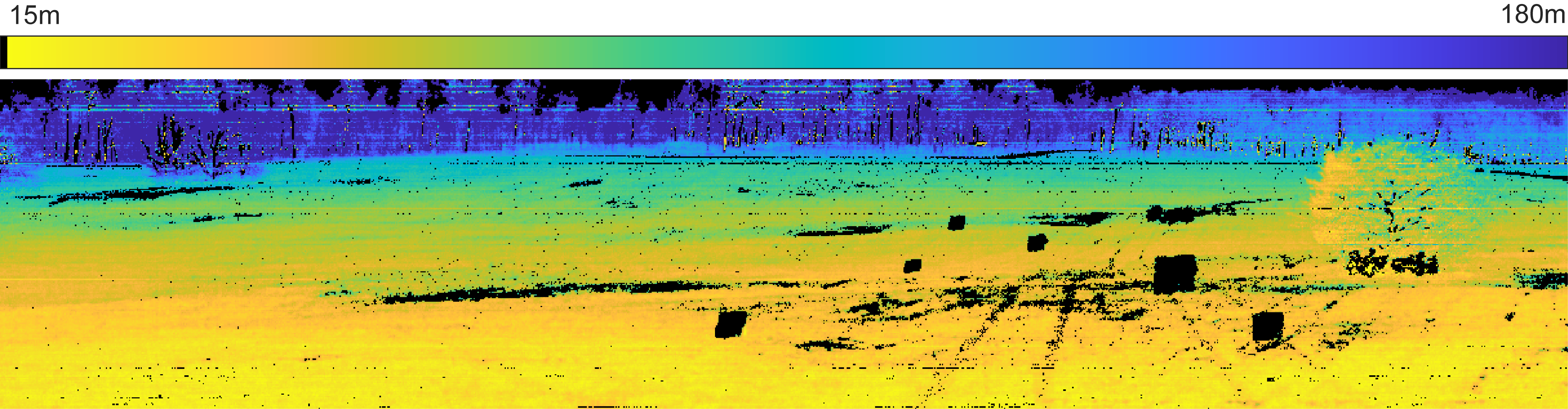}\label{fig:Ozone-04}} \hfill
    \caption{Detection of overestimated pixels due to downwelling radiance.
    \protect\subref{fig:Ozone-01} Absolute difference of measurements between 9.50\,\si{\micro\meter} and 9.58\,\si{\micro\meter}, showing a measure of the downwelling radiance contribution.
    \protect\subref{fig:Ozone-02} Binary mask by thresholding the ozone difference image.
    \protect\subref{fig:Ozone-03} Range estimate for all pixels.
    \protect\subref{fig:Ozone-04} Range estimate for reliable (not affected by downwelling) pixels.
    The contribution of downwelling radiance is different for each pixel depending on the reflectivity ($1-\emiss(\lambda)$), and orientation captured with the solid angle $\Omega$ in~\eqref{eq:ReflectionContribution}.
    For natural scenes, mostly composed of vegetation, the contribution from downwelling radiance is small due to high emissivity.}
    \label{fig:ozone_comparison}
\end{figure}

\begin{figure*}
    \centering
    \subfloat[Grass patch around 45\,\si{\meter}]{\includegraphics[width=0.33\textwidth]{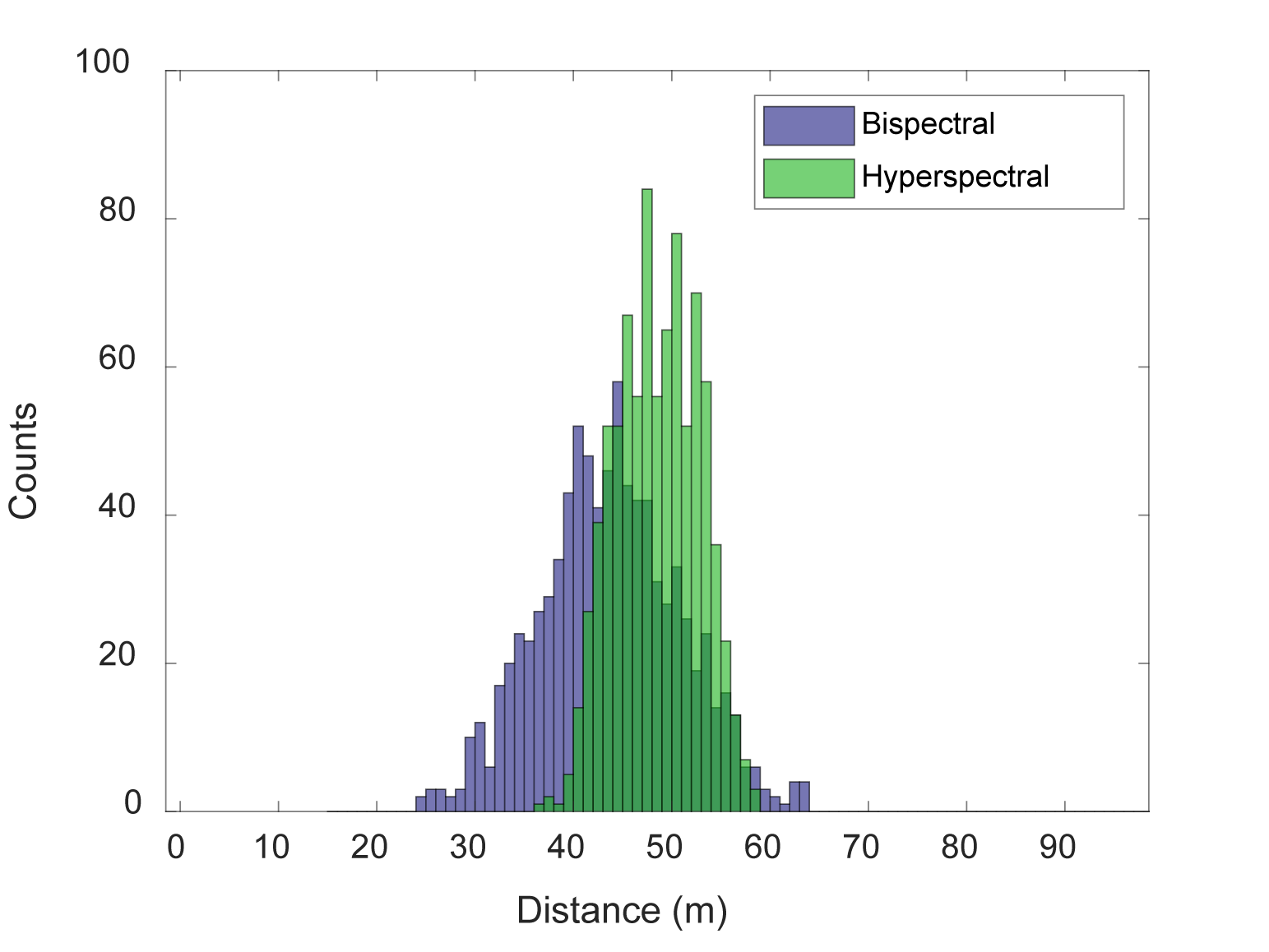}\label{fig:HistogramFarBacgroundForest}} \hfill
    \subfloat[Foreground tree patch around 70\,\si{\meter}]{\includegraphics[width=0.33\textwidth]{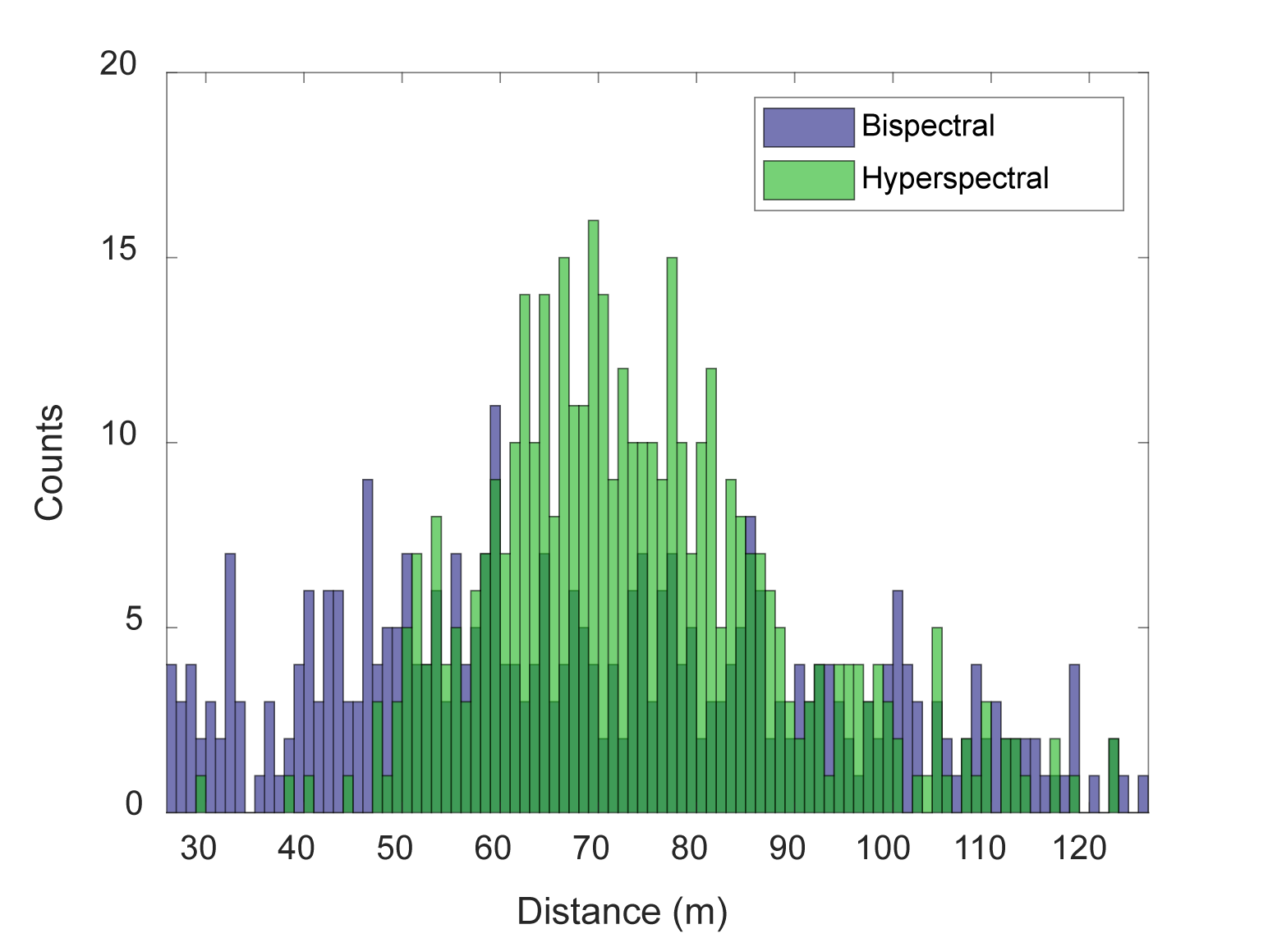}\label{fig:ForegroundTree}} \hfill
    \subfloat[Background forest patch around 160\,\si{\meter}]{\includegraphics[width=0.33\textwidth]{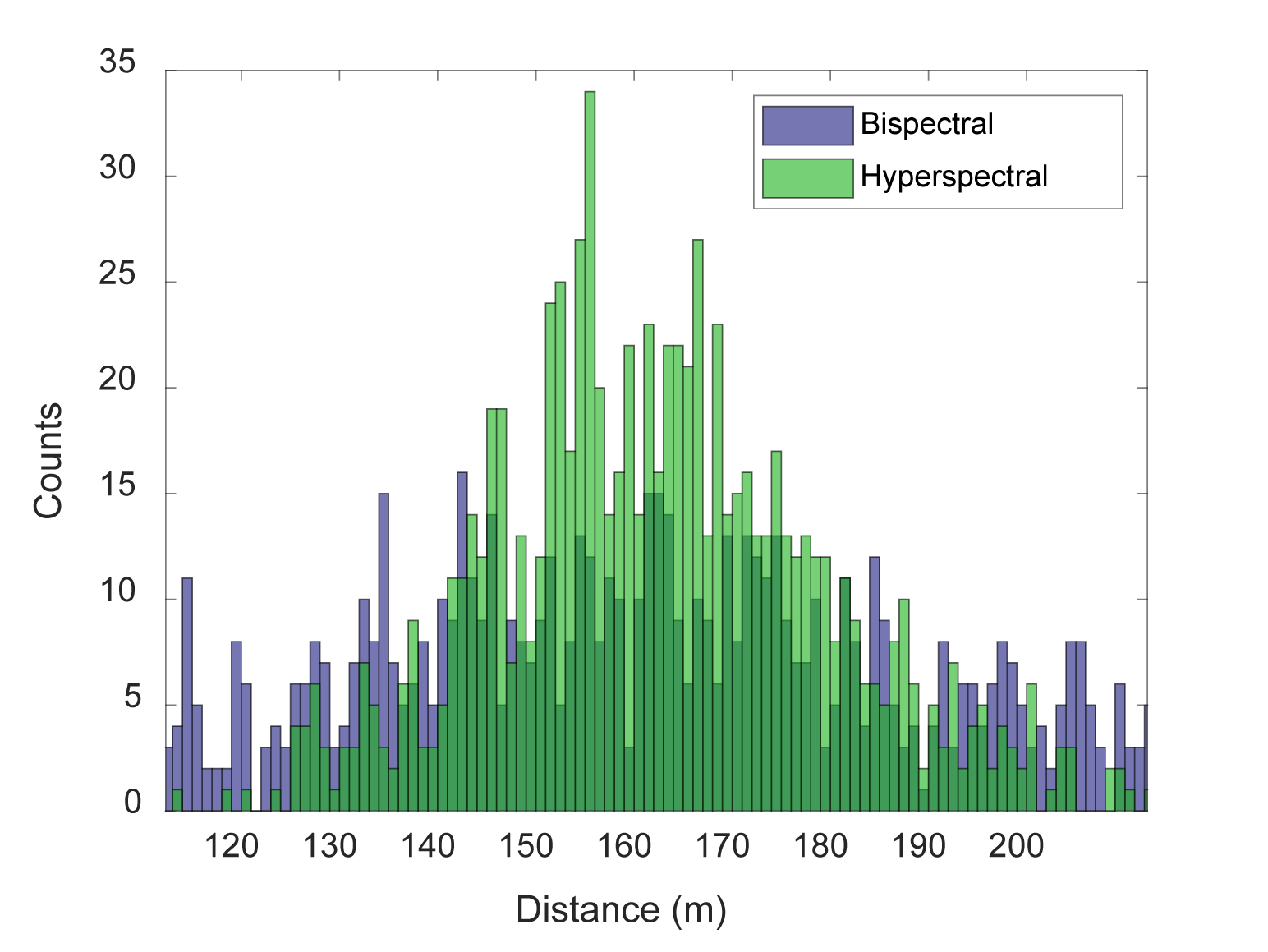}\label{fig:HistogramMidGrass}}
    \caption{Experimental results of bispectral (purple) and hyperspectral (green) ranging methods for three $20 \times 40$ patches within which the distance is presumably approximately constant.
    Unlike the simulated results, we cannot show error as the ground truth is only approximate from the lidar.
    Instead, the means and the standard deviations of the histograms are compared in Table~\ref{tab:Experimental_table}.
    }
    \label{fig:ExperimentalHistograms}
\end{figure*}

\begin{figure}
    \centering
    {\includegraphics[width=0.5\textwidth]{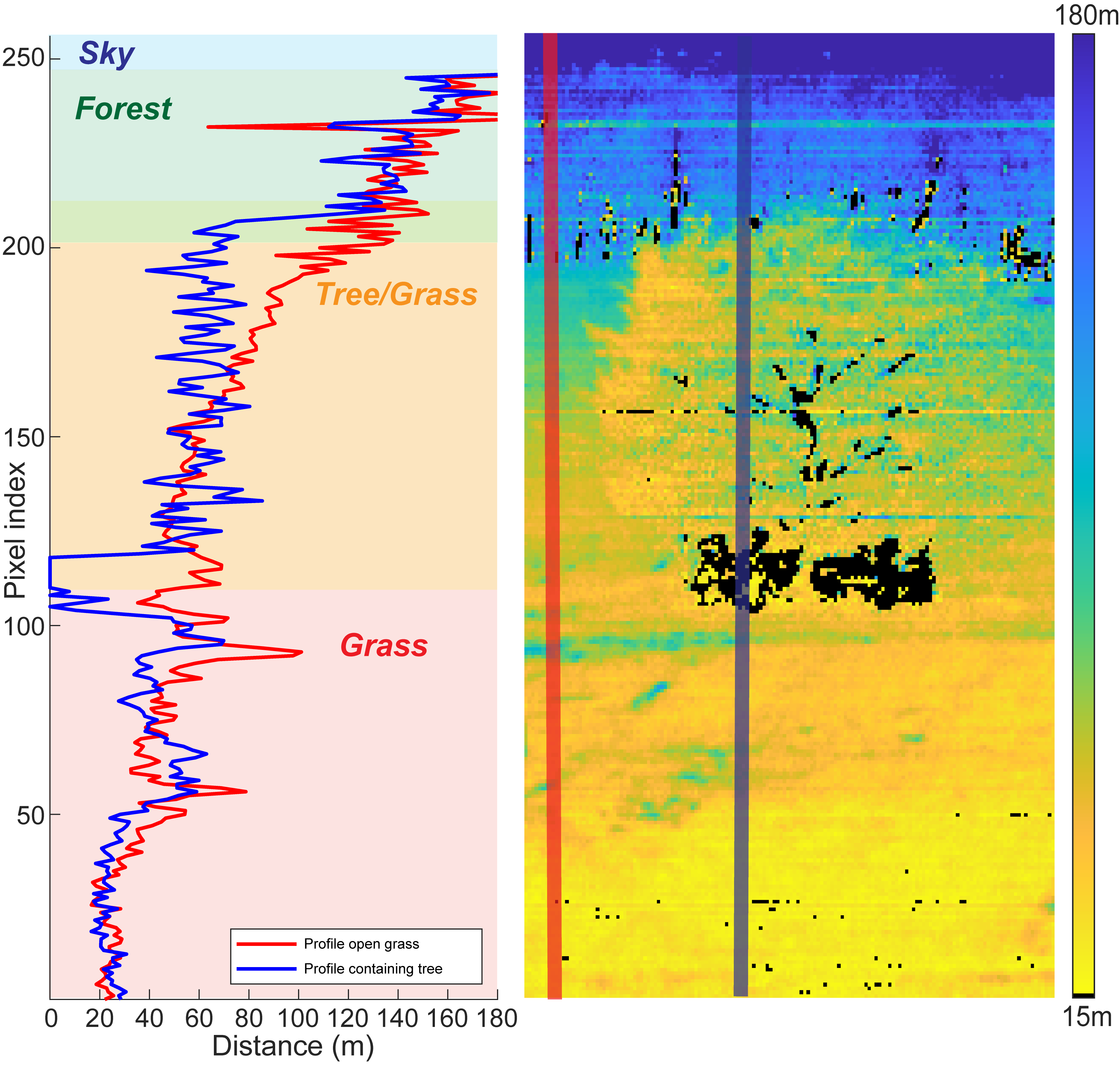}}
    \caption{Extracted range profile of two vertical lines, one containing the tree (blue) and one at open grass terrain (red) as indicated in the figure on right.
    On the left,
    shaded areas roughly represent different parts of the scene:
    sky (light blue),
    background forest (green),
    foreground tree/grass (yellow),
    and grass (red).
    The tree is clearly distinguishable from background forest with a sharp transition from 70\,\si{\meter} to 150\,\si{\meter} on the blue curve around pixel index 210.
    The rolling grassy terrain is also distinguishable from the tree, since from pixel index 100 to 200 there is a smooth transition on the red curve from around 60\,\si{\meter} to 130\,\si{\meter} whereas the blue curve is roughly constant around 70\,\si{\meter}.
    }
    \label{fig:ExperimentalVerticalSlice}
\end{figure}

\begin{figure}
    \centering
    \subfloat[Nearby (30\,\si{\meter}) grass area]{\includegraphics[width=0.4\textwidth]{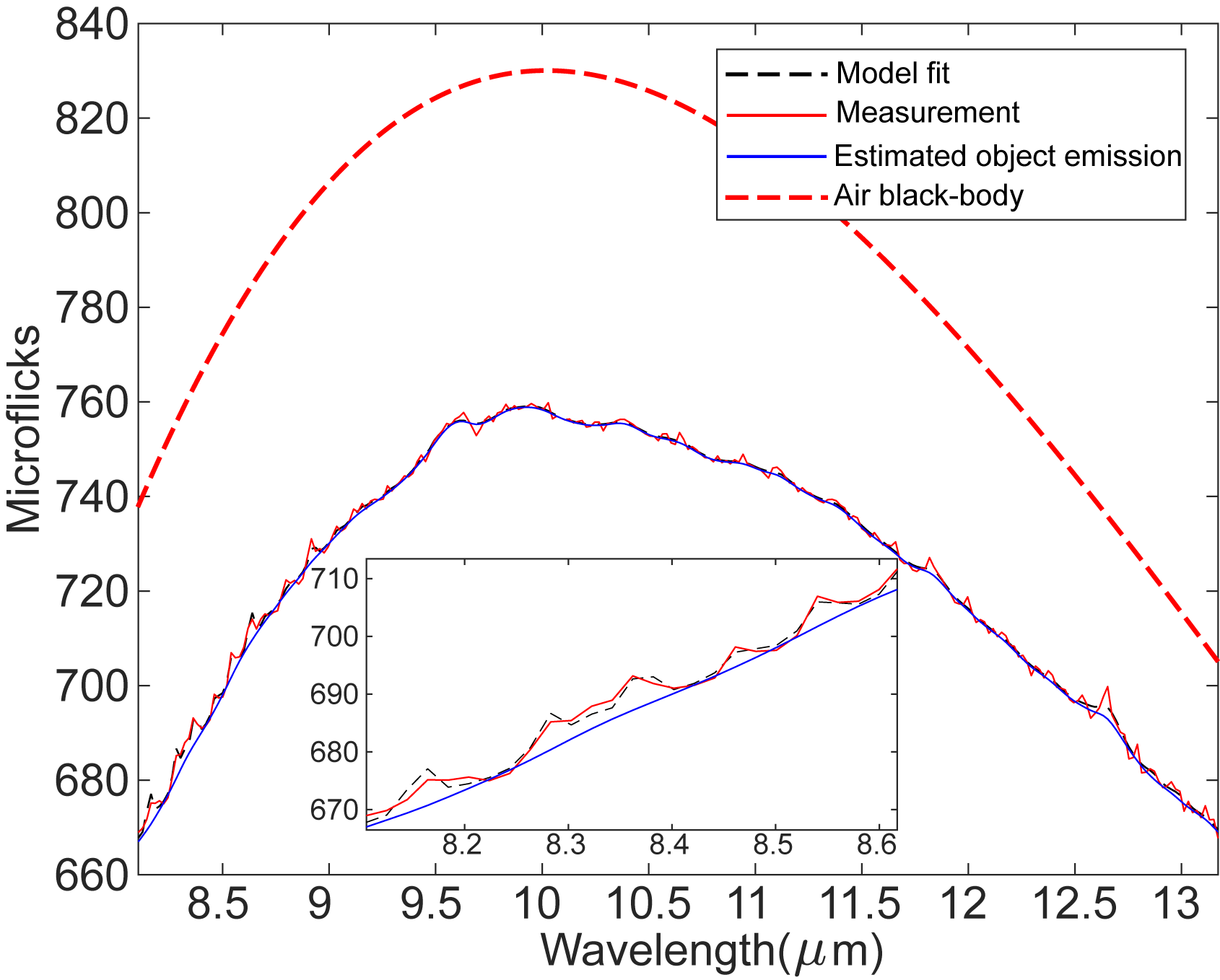}\label{fig:Measurement_fit_Nearby}} \\
    \subfloat[Distant (85\,\si{\meter}) grass area]{\includegraphics[width=0.4\textwidth]{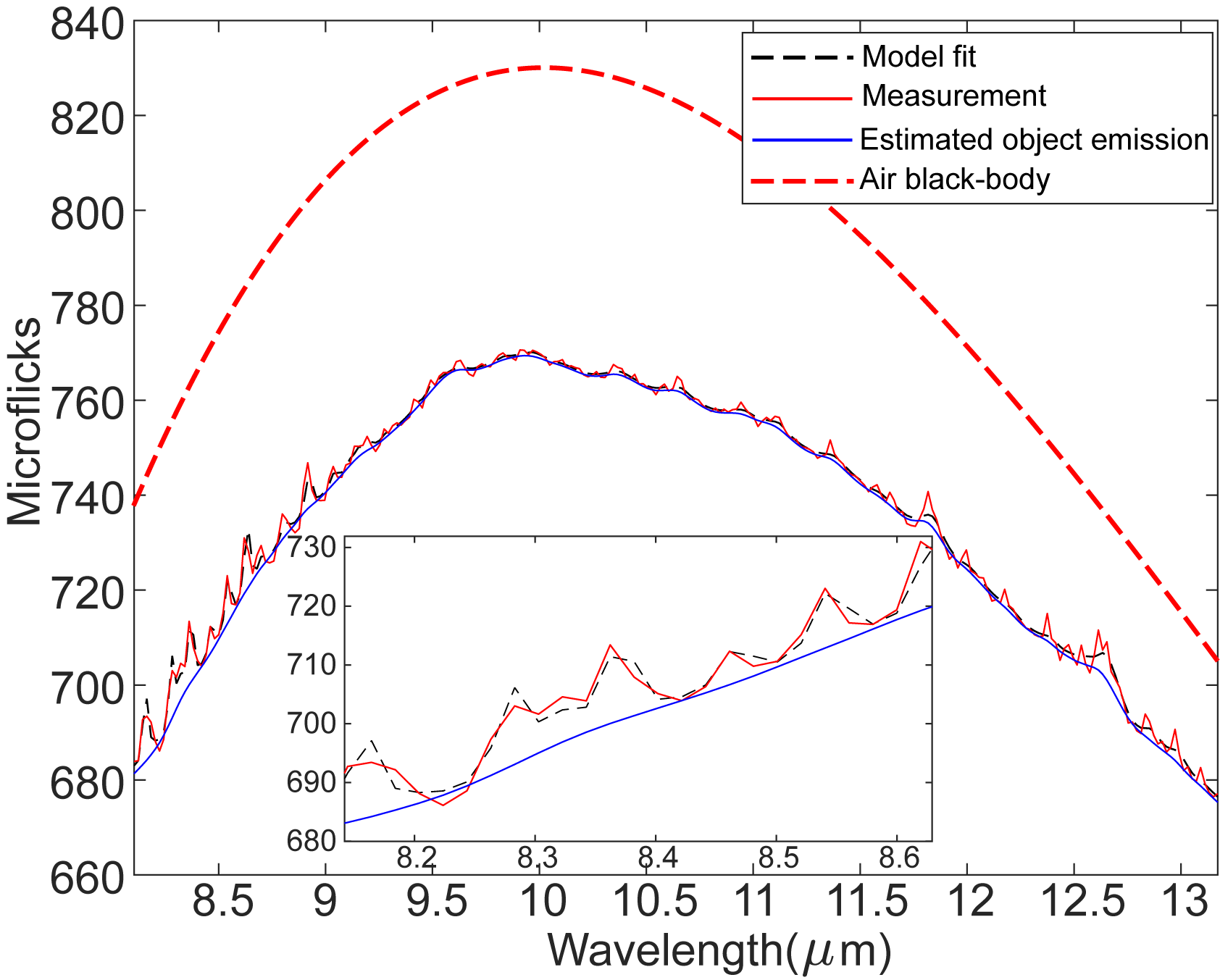}\label{fig:Measurement_fit_Far}} 
    \caption{Model fit to the measurements:
    \protect\subref{fig:Measurement_fit_Nearby} is for a nearby (30\,\si{\meter}) grass area; and
    \protect\subref{fig:Measurement_fit_Far} is for a distant (85\,\si{\meter}) grass area.
    }
    \label{fig:ModelFit}
\end{figure}

\subsubsection{Detection of Significant Downwelling Radiation Pixels}
\label{sec:experimental_downwelling_detection}
As discussed in Section~\ref{sec:forward}, our hyperspectral ranging method is based on
the forward model \eqref{eq:ForwardModel},
which neglects reflected light.
While any reflected light creates a model mismatch,
reflected sky radiation,
otherwise known as downwelling radiance~\cite{kirkland2002thermal},
is especially significant for our ranging problem.
It contains spectral variations characteristic of long propagation distances
and hence leads to overestimations of range.
We seek a method to detect a large component of reflected downwelling radiance in a pixel so that we can classify the pixel's range estimate as unreliable.

We propose to detect reflected downwelling radiance by exploiting the ozone absorption line around 9.6\,\si{\micro\meter}.
Because the ozone concentration at ground level is typically negligible~\cite{coesa1976standard}, signal variation
around this wavelength are attributed to downwelling radiance
that has traveled through the ozone layer.
Fig.~\ref{fig:ozone_comparison}\subref{fig:Ozone-01} shows the absolute difference of measurements between 9.50\,\si{\micro\meter} and 9.58\,\si{\micro\meter}, around the ozone absorption line.
The sky pixels that are in direct line of sight (yellow pixels at the top edge of the scene) show large values for this difference.
Most other pixels show a small value for this difference.
The value of this difference at each pixel is based on the contribution of the downwelling radiance to the measurements, which depends on the reflectivity, shape, and orientation of the object.

For highly emissive objects, the contribution from downwelling is negligible.
Most of the scene is composed of vegetation, which is known to have high emissivity (low reflectivity), and thus is less impacted by downwelling radiation.  
For low emissivity (high reflectivity) objects, the contribution from the downwelling radiance can become significant, leading to overestimates of distance.
One such case can be seen in the checkerboard panels, which are highly reflective.
We classify the pixels that have significant downwelling radiance contribution by thresholding the ozone difference image.
With a threshold of 4 microflicks,
7\% of the pixels are classified as having unreliable estimates, as shown in
Fig.~\ref{fig:ozone_comparison}\subref{fig:Ozone-02}.
Fig.~\ref{fig:ozone_comparison}\subref{fig:Ozone-03} shows the range estimate for all pixels, and Fig.~\ref{fig:ozone_comparison}\subref{fig:Ozone-04} shows the range estimates excluding pixels that have significant downwelling radiance contribution using the mask.
Most of the overestimated pixels, such as the checkerboard panels and grass areas, are detected using the ozone difference image.

For highly reflective regions, neglecting downwelling radiance in the inversion can also propagate into emissivity estimates.
In~Fig.\ref{fig:EmissivityClusters}, the sky pixels assigned to Cluster 3 and the reflective panel pixels assigned to Cluster 4 each show a strong feature around 9.6\,\si{\micro\meter} that is consistent with ozone emission from the upper atmosphere~\cite{kirkland2002thermal}.
A more sophisticated model that included reflections and downwelling would be less susceptible to this type of errors.
The parts of the scene where the relative temperature is low (background forest and foreground tree) are noisy compared to other parts that have high relative temperature (grassy terrain).

\subsubsection{Evaluation}
\label{sec:experimental_evaluation}
We cannot quantitatively evaluate the errors of the distance estimates in Fig.~\ref{fig:ExperimentalResultsP5S1}\subref{fig:Experimental_results_Depth_map} because we do not have exact pixel-by-pixel matching between hyperspectral images and lidar data.
Instead, we calculate histograms of estimated distances for $20 \times 40$ patches of the scene within which the true distance is presumably approximately constant.
Fig.~\ref{fig:ExperimentalHistograms} shows three such histograms for each method.
The grass patch (Fig.~\ref{fig:HistogramMidGrass}), estimated around 45\,\si{\meter}, shows the best performance (least variance across the patch).
The foreground tree (Fig.~\ref{fig:ForegroundTree}), estimated around 70\,\si{\meter}, has low relative temperature and shows the worst performance.
The background forest patch (Fig.~\ref{fig:HistogramFarBacgroundForest}) is estimated around 160\,\si{\meter}; the performance here is worse than for the grassy area, but slightly better than for the foreground tree.
Table~\ref{tab:Experimental_table} shows the means and standard deviations of the histograms.

\begin{table}
\caption{Means and standard deviations of experimental results on $20 \times 40$ patches; histograms shown in Fig.~\ref{fig:ExperimentalHistograms}.
}
\label{tab:Experimental_table}
\centering
\scalebox{0.92}{
\begin{tabular}{@{}rr@{\,\,}c@{\,\,}rr@{\,\,}c@{\,\,}rr@{\,\,}c@{\,\,}r@{}}
\toprule
Patch &
\multicolumn{3}{r}{Grass} &
\multicolumn{3}{r}{Foreground tree} &
\multicolumn{3}{r@{}}{Background forest} \\
\midrule
Bispectral    &  43.7\,\si{\meter} & $\pm$ &  7.2\,\si{\meter}
              &  67.8\,\si{\meter} & $\pm$ & 53.8\,\si{\meter}
              & 166.2\,\si{\meter} & $\pm$ & 41.0\,\si{\meter}\\
Hyperspectral &  48.5\,\si{\meter} & $\pm$ & {\bf 4.2}\,\si{\meter}
              &  76.0\,\si{\meter} & $\pm$ & {\bf 21.2}\,\si{\meter}
              & 162.3\,\si{\meter} & $\pm$ & {\bf 17.8}\,\si{\meter} \\
\bottomrule
\end{tabular}
}
\end{table}

To further analyze the results,
distance profiles of two vertical lines, estimated via the hyperspectral method, are plotted in Fig.~\ref{fig:ExperimentalVerticalSlice}.
One vertical line contains the foreground tree (blue) and the other vertical line does not (red).
The top pixels represent the sky, shaded blue. As the pixel index decreases, the field-of-view moves to nearer objects.
The background forest and foreground tree/grass pixels are shaded green and yellow, respectively.
The grass pixels are shaded red.
Comparing the two vertical distance profiles in solid red and blue, the top pixels corresponding to sky are estimated higher than 180\,\si{\meter} as expected.
The estimations for the background forest area match between the two vertical distance profiles around 160\,\si{\meter}.
Focusing near pixel 200, the tree is clearly distinguishable from the background forest, with a sharp transition from 70\,\si{\meter} to 150\,\si{\meter} on the blue curve.
The rolling grassy terrain is also distinguishable from the tree, since from pixel index 100 to 200 there is a smooth transition on the red curve from around 60\,\si{\meter} to 130\,\si{\meter}, whereas the blue curve is roughly constant around 70\,\si{\meter}.
This latter result is in good agreement with the lidar measurements, which indicate this tree is approximately 65\,\si{\meter} away from the sensors.

Fig.~\ref{fig:ModelFit} shows the measurements and model fits for two individual pixels in the grassy area.
One is for a nearby pixel, estimated at around 30\,\si{\meter}, and the other is for a distant grass pixel estimated around 85\,\si{\meter} from the sensor.
The measurements are represented in red, the model fit in dashed-black, the emission fit in blue, and the air black-body in dashed-red.
The emissions from these objects are similar, as can be validated from the emission fits.
The distant grass measurements in Fig.~\ref{fig:ModelFit}\subref{fig:Measurement_fit_Far} have significantly sharper peaks at the absorption bands compared to nearby grass pixels in Fig.~\ref{fig:ModelFit}\subref{fig:Measurement_fit_Nearby}.
The model fit in each case is excellent at the absorption bands, whereas the emission fit is smooth over the spectrum.
This further suggests that the sharp peaks are due to atmospheric absorption and emission.

\section{Discussion}
\label{sec:Conclusion}
In this paper, we showed range imaging methods using atmospheric absorption from ambient thermal radiation.
In natural scenes, the temperature variation is typically limited to a few degrees.
Therefore, it is crucial to account for the air emission to avoid model mismatch.
We extended the bispectral range estimator to account for air emission.
Results suggest that not accounting for air emission causes extremely poor range estimates.
Furthermore, we showed that the main limitation for ranging is the temperature difference between the air and the object.
As the temperature difference decreases, the measurements are less sensitive to distance.
This is shown both experimentally, via histograms, and theoretically by Fisher information analysis.
For low-contrast scenarios, such as in natural scenes, it is helpful to include as many absorption lines as possible to alleviate the effects of noise.

Fortunately, hyperspectral measurements provide channels with a variety of transmission levels.
Depending on the transmission level, each channel can be informative about a different parameter.
Saturated absorptive channels ($\tau(\lambda;d) \approx 0$) are useful for air temperature estimates, as the measurements approach the air black-body curve.
On the other hand, highly transmissive channels ($\tau(\lambda;d) \approx 1$) are useful for object temperature and emissivity estimates.
Other channels are useful for ranging.
We showed the trade-off in atmospheric attenuation levels:
If the attenuation level is very high, the emitted radiance does not reach the sensor; conversely, if the attenuation level is very low, there is not enough absorption for ranging.
Our Fisher information analysis suggests
that the optimal attenuation is $1/e$ or 4.3\,\si{\dB}.

Through both simulations and the processing of experimental data, we demonstrated a hyperspectral ranging method that jointly estimates object temperature, emissivity profile, and range.
The joint estimation is crucial for the accuracy of ranging while processing a wider band as both object emissivity and atmospheric transmission are wavelength dependent.
The problem is underdetermined and requires some assumptions to be solvable.
For this, we exploited the smooth spectral variations of solid objects compared to the sharp absorption lines of the atmosphere to find good solutions.
Processing a wider band of the spectrum is valuable for many reasons.
First, accounting for $100\%$ of the available information from ambient radiation, all bands should be considered.
Second, the attenuation level is dependent on the weather conditions and thus not controllable by the user;
it is important to span a large spectral range to include different attenuation coefficients for optimal ranging at a variety of distances.
Last, the hyperspectral estimation is not only important to reduce the effect of noise, but it is also important to extract the emissivity profile of objects.
To show the value of the estimated emissivity profiles, $k$-means clustering was performed.
The results suggest that materials can be classified from their estimated emissivity profiles.
Furthermore, we also discuss how the reflected downwelling radiance can cause overestimations for highly reflective objects, as it includes a reflected component that traveled a much longer path through the atmosphere before reaching the sensor.
To address this, we proposed a method to identify pixels significantly impacted by downwelling radiance by leveraging ozone absorption features.
\ifCLASSOPTIONcompsoc
  \section*{Acknowledgments}
\else
  \section*{Acknowledgment}
\fi

Data provided by the U.S. Army Night Vision and Electronic Sensors Directorate 
and Johns Hopkins University Applied Physics Laboratory.

\ifCLASSOPTIONcaptionsoff
  \newpage
\fi

\bibliographystyle{IEEEtran}
\input{Main_journal.bbl}

\begin{IEEEbiography}[{\includegraphics[width=1in,height=1.25in,clip,keepaspectratio]{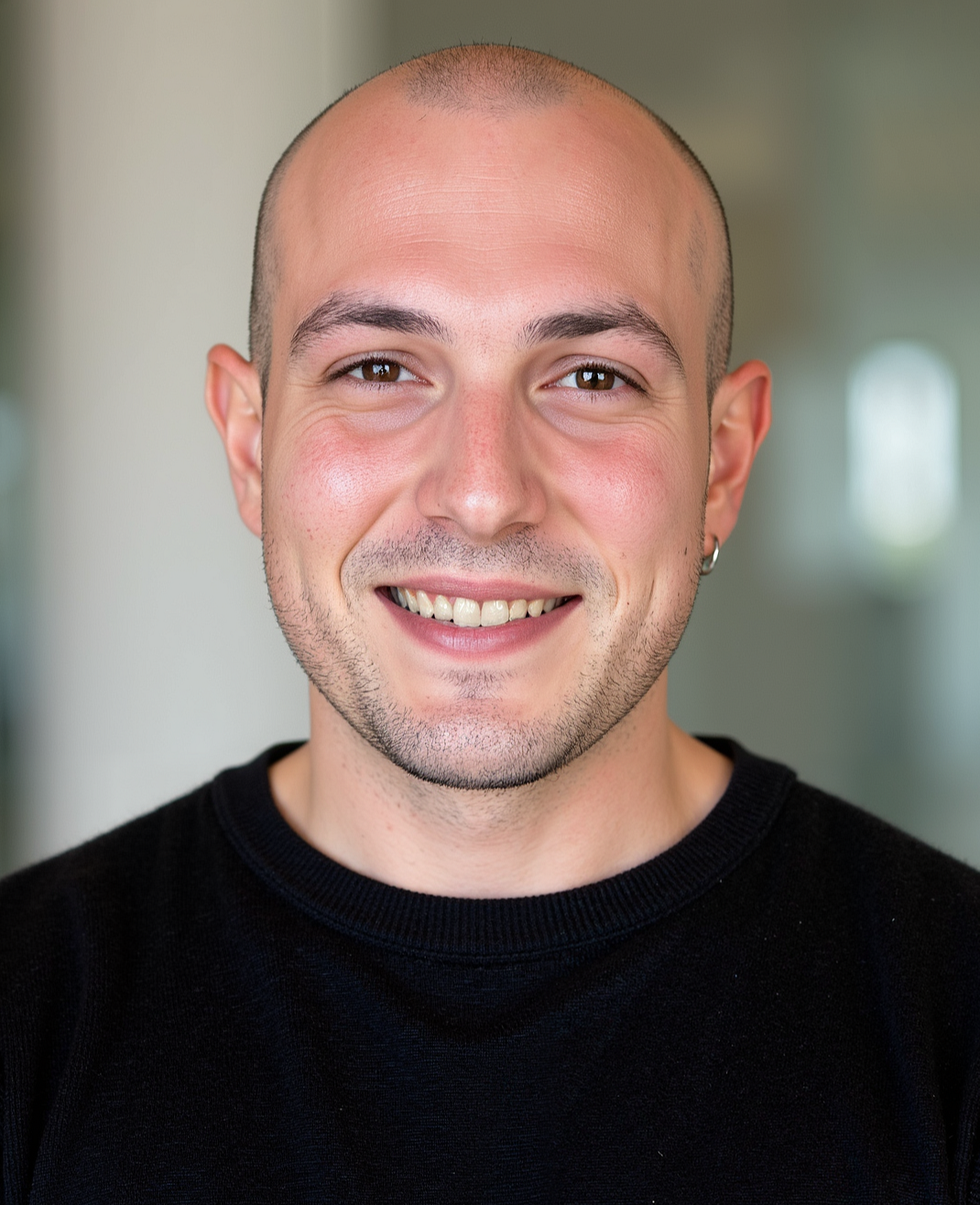}}]{Unay Dorken Gallastegi}
received the B.S. degree from the Department of Electric and Electronics Engineering, Bilkent University, Ankara, Turkey, in 2019\@.
He is currently working toward the Ph.D. degree in Electrical and Computer Engineering at Boston University, Boston, MA.
He joined Meta, Redmond, WA, in 2022 as research scientist intern.
His research interests include computational imaging, hyperspectral imaging, 3D reconstruction, and image processing.
\end{IEEEbiography}

\begin{IEEEbiography}[{\includegraphics[width=1in,height=1.25in,clip,keepaspectratio]{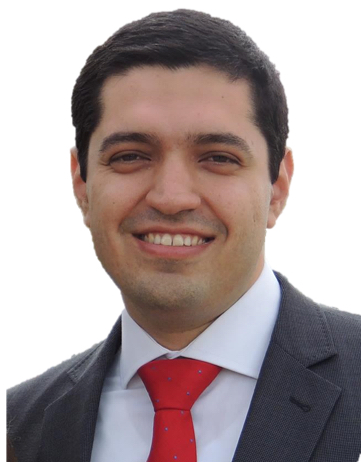}}]{Hoover Rueda-Chacon}
(S'12--M'18) received the B.Sc. and M.Sc. degrees in Computer Science from the Universidad Industrial de Santander, Colombia, in 2009 and 2012, respectively, and the M.Sc. and Ph.D. degrees in Electrical and Computer Engineering from the University of Delaware, Newark, DE, USA, in 2015 and 2017, respectively, sponsored by a Fulbright-Colciencias scholarship.
From 2021 to 2022, he was a Postdoctoral Associate with the Department of Electrical and Computer Engineering at Boston University, USA\@.
At present, he is an Assistant Professor at Universidad Industrial de Santander, Colombia.
His main research interests include computational imaging, hyperspectral imaging, single-photon imaging, high dimensional signal processing and optimization algorithms.
\end{IEEEbiography}

\begin{IEEEbiography}[{\includegraphics[width=1in,height=1.25in,clip,keepaspectratio]{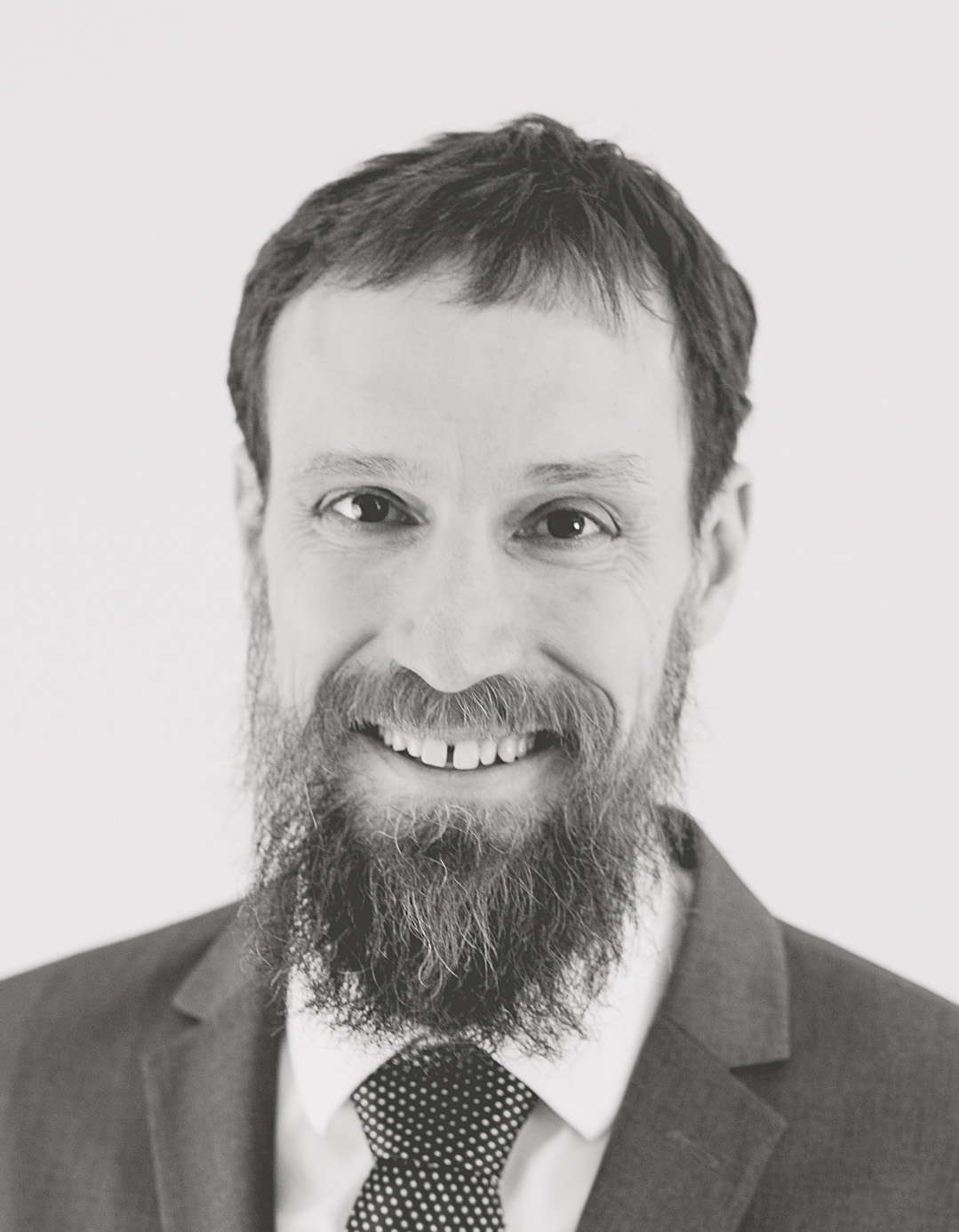}}]{Martin J. Stevens}
received the B.S. degree in physics from the University of Minnesota in 1996\@.
He received the Ph.D. degree in electrical and computer engineering from the University of Iowa in 2004\@.
From 2004 to 2007 he was a postdoctoral scholar at the National Institute of Standards and Technology (NIST) in Boulder, CO, where he was awarded a National Research Council Fellowship.
Since 2007 he has been a member of the technical staff at NIST, where he has been awarded Gold, Bronze and Silver Medals from the Department of Commerce.
Dr.\ Stevens is a member of Optica, and served on the organizing committee for the CLEO conference from 2017 to 2021\@.
He was co-organizer of the Single Photon Workshop in 2017, and co-organizer of the NIH Virtual Workshop: Near-term Applications of Quantum Sensing Technologies in Biomedical Sciences in 2023.
\end{IEEEbiography}

\begin{IEEEbiography}[{\includegraphics[width=1in,height=1.25in,clip,keepaspectratio]{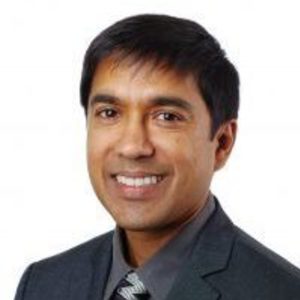}}]{Vivek K Goyal}
(S'92--M'98--SM'03--F'14) received the B.S. degree in mathematics and
the B.S.E. degree in electrical engineering from the University of Iowa,
where he received the John Briggs Memorial Award for the top undergraduate
across all colleges.
He received the M.S. and Ph.D. degrees in electrical engineering from
the University of California, Berkeley, where he received the Eliahu Jury Award
for outstanding achievement in systems,
communications, control, or signal processing.

He was a Member of Technical Staff in the Mathematics
of Communications Research Department of Bell Laboratories,
Lucent Technologies, 1998--2001; and
a Senior Research Engineer for Digital Fountain, Inc., 2001--2003\@.
He was with the Massachusetts Institute of Technology 2004--2013,
where he was the Esther and Harold E. Edgerton Associate Professor
of Electrical Engineering.
He was an adviser to 3dim Tech, Inc.\ (winner of the
2013 MIT \$100K Entrepreneurship Competition Launch Contest Grand Prize),
and was subsequently with Nest, an Alphabet company, 2014--2017\@.

Dr.\ Goyal is a member of
Phi Beta Kappa and Tau Beta Pi.
He was awarded
the 2002 IEEE Signal Processing Society (SPS) Magazine Award,
the 2014 IEEE Int.\ Conf.\ Image Processing Best Paper Award,
and the IEEE SPS Best Paper Award in 2017 and 2019\@.
Work he supervised won student best paper awards at
the IEEE Data Compression Conf.\ in 2006 and 2011,
the 2012 IEEE Sensor Array and Multichannel Signal Processing Workshop, and
the 2018 IEEE Int.\ Conf.\ Image Processing,
as well as
the 2020 IEEE SPS Young Author Best Paper Award and
the 2021 IEEE SPS Best PhD Dissertation Award.
He was
a Co-chair of the SPIE Wavelets and Sparsity
conference series 2006--2016\@.
He current serves on the Editorial Boards of
{\sc IEEE Trans.\ Computational Imaging},
{\sc SIAM J. Imaging Sciences},
and
\emph{Foundations and Trends in Signal Processing}.
He is a Fellow of AAAS, the Guggenheim Foundation, and Optica and a co-author of \emph{Foundations of Signal Processing}
(Cambridge University Press, 2014).
\end{IEEEbiography}

\end{document}